%% file: format.tex
\documentclass[11pt,letterpaper]{article}
\usepackage{cogsys}
\usepackage[T1]{fontenc}
\usepackage{times}
\usepackage[pdftex]{graphicx} % use this when importing PDF files
\usepackage{xcolor}
\usepackage{subfig}
\usepackage{wrapfig}
\usepackage{amsmath}

\usepackage{natbib}
\cogsysheading{X}{20XX}{1-6}{X/20XX}{X/20XX}
\input{cup-macros.tex}

\ShortHeadings{Detecting and Accommodating Novel Types and Concepts}
              {S.\ Ghaffari and N.\ Krishnaswamy}

\begin{document} 

\title{Detecting and Accommodating Novel Types and Concepts in an Embodied Simulation Environment}
 
\author{Sadaf Ghaffari}{sadafgh@colostate.edu}
\author{Nikhil Krishnaswamy}{nkrishna@colostate.edu}
\address{Situated Grounding and Natural Language Lab, Department of Computer Science, Colorado State University, 
         Fort Collins, CO 80523 USA}
\vskip 0.2in
 
\begin{abstract}In this paper, we present methods for two types of metacognitive tasks in an AI system: rapidly expanding a neural classification model to accommodate a new category of object, and recognizing when a novel object type is observed instead of misclassifying the observation as a known class. Our methods take numerical data drawn from an embodied simulation environment, which describes the motion and properties of objects when interacted with, and we demonstrate that this type of representation is important for the success of novel type detection. We present a suite of experiments in rapidly accommodating the introduction of new categories and concepts and in novel type detection, and an architecture to integrate the two in an interactive system.
\end{abstract}

\vspace*{-2mm}
\section{Introduction}
\label{sec:intro}
\vspace*{-2mm}

Humans are efficient at seeking out experiences that are maximally informative about their environment \citep{markant2014better,najemnik2008eye,nelson2010experience,renninger2007look,schulz2007serious}.  We explore the physical world to practice skills, test hypotheses, learn object affordances, etc. \citep{caligiore2008using,gopnik1997words,gopnik2010babies,gopnik2012scientific,gottlieb2018towards,neftci2019reinforcement,piaget1963attainment,piaget2008psychology,son2006metacognitive}. Young children, in particular, can rapidly expand their vocabulary of concepts with few or no examples, and generalize from previous to new experiences \citep{clark2006language,colung2003emergence,vlach2012fast}.

Meanwhile, artificial neural networks require large numbers of samples to train. It may take 5-8 layers of artificial neurons to approximate a single cortical neuron \citep{beniaguev2021single}. Common few-, one-, or zero-shot learning approaches in AI provide at best a rough simulacrum of human learning and generalization \citep{knudsen1994supervised,niv2009reinforcement,zador2019critique}. Recent successes in few-shot learning in end-to-end deep neural systems still require extensive pre-training and fine-tuning, often on special hardware \citep{brown2020language}, or specific task formulation \citep{schick2020s}. They do not easily or organically expand to accommodate new concepts.

%Explicit representation has been disfavored in modern AI, but typical neural networks that learn implicit representations are treated as passive recipients of data, with the question of context- and situation-sensitive grounding treated as something of an inconvenience \citep{chen2015deepdriving}. There is little attention paid to letting the current state of the world, as opposed to reams of pre-existing data, be "its own model" a la \cite{brooks1991intelligence}.

In this paper, we present an investigation into the ability of machine learning systems to engage in certain metacognitive behaviors pertaining to the detection and acquisition of new concepts.  In the context of this paper, we take a ``metacognitive'' process to be one that requires the system to be aware of what it does and doesn't know. This is, of course, just one factor that makes an agent ``metacognitive'' \citep{meichenbaum1985teaching}. Additionally, to make our problem tractable, we focus on the particular domain of object types, and more abstract concepts that inhere across multiple object types, such as the opposition between ``flatness'' and ``roundness.''  We model the concept acquisition problem as one of an agent engaged in open-ended play with simple objects, as might be the case for an infant or toddler who starts off interacting with only a few classes of simple objects before other comparatively more complex objects are introduced.  In particular, we address two questions: 1) How can a model grow to accommodate novel object types, and 2) How can an agent automatically detect that a new object type has been introduced into the environment such that it knows to update its model?

\vspace*{-4mm}
\section{Related Work}
\label{sec:related}
\vspace*{-2mm}

This work has many antecedents in earlier AI, particularly early explorations into multi-task learning \citep{caruna1993multitask} and knowledge transfer between neural networks \citep{pratt1991direct}.

Object recognition and classification is of course a well-traveled area in AI, but the AI approaches also have antecedents in the cognitive science community.  Among many others, \cite{riesenhuber2000models} presented models of computational object recognition inspired by processes in the human visual cortex, \cite{oliva2007role} motivated development on pre-neural network computer vision systems through examining human use of contextual cues in object recognition, and \cite{dicarlo2007untangling} drew on both neurophysiology and computation to examine the brain mechanisms that allow for rapid object recognition under multiple circumstances.  \cite{piloto2022intuitive} is one of a number of recent approaches that explore similar questions in vision, but uses an order of magnitude more samples than our approach, and does not address the contribution of interacting with objects directly.

Among approaches where the interaction between agent and environment are central, \cite{nolfi2005category}, drawing on ``embodied cognitive science'' \citep{scheier1999embodied}, proposed a theory of category formation based on the results of interacting with the environment in simple tasks.  \cite{mohan2012acquiring} and \cite{frasca2018one} are examples of work from the cognitive systems community that address language grounding and acquisition in situated environments. \cite{bar2006functional} used simulation to classify objects based on their functional properties, but did not look at identifying when a novel class has been introduced.

\cite{fitzgerald2021abstraction} consider similar questions, where they begin with a model of a task and then attempt to transfer that model to a related task that is identified through perception, while also recognizing failure.  Metacognition, even in the limited scope as we have defined it herein, is supervenient upon such capabilities.

\vspace*{-4mm}
\section{Environment and Data Collection}
\label{sec:environment-data}
\vspace*{-2mm}

\begin{figure}[h!]
    \centering
    \includegraphics[height=.7in]{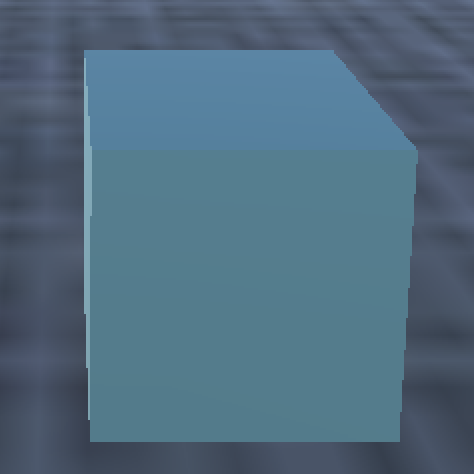}
    \includegraphics[height=.7in]{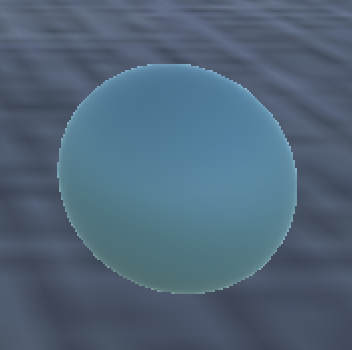}
    \includegraphics[height=.7in]{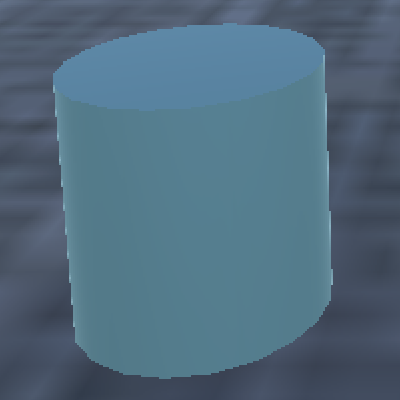}
    \includegraphics[height=.7in]{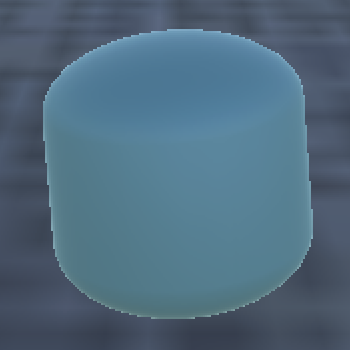}
    \includegraphics[height=.7in]{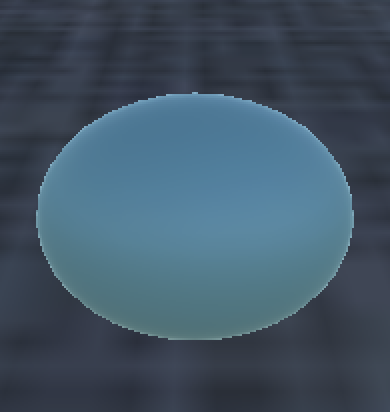}
    \includegraphics[height=.7in]{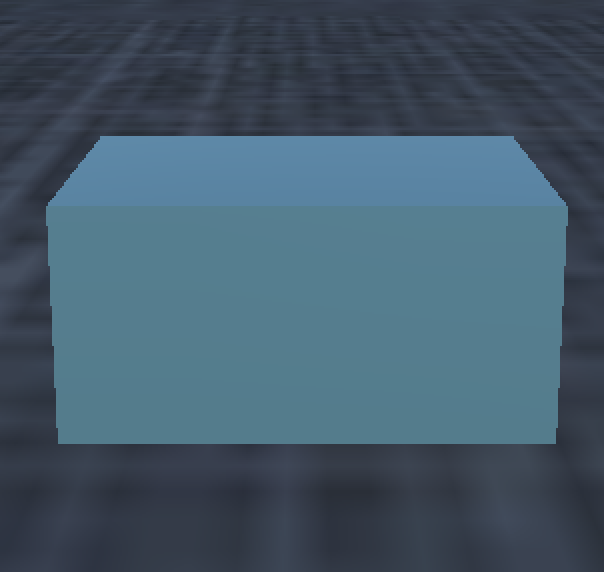}
    \includegraphics[height=.7in]{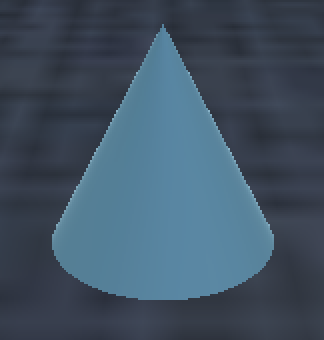}
    \includegraphics[height=.7in]{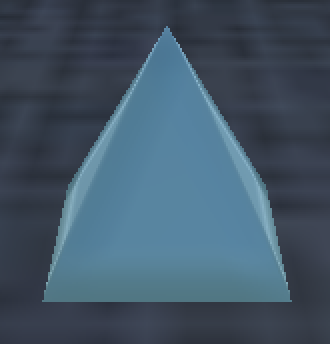}
    \vspace*{-3mm}
    \caption{Possible theme objects, shown in default orientation: {\it cube}, {\it sphere}, {\it cylinder}, {\it capsule}, {\it egg}, {\it rectangular prism}, {\it cone}, {\it pyramid}.  Also included is a {\it small cube} (not shown).}
    \vspace*{-3mm}
    \label{fig:objects}
\end{figure}

To create an environment where an agent can explore object properties through interaction, we use the VoxWorld platform for interactive agents \citep{krishnaswamy-EtAl:2022:LREC}.  We create an environment where an agent is presented with two objects: one cube and one instance of another class, known as the {\it theme object} (see Fig.~\ref{fig:objects}). The objects all differ from each other in their geometric features---particularly where they have {\it round} sides and {\it flat} sides---and the object set is broadly inspired by geometric children's toys. Further examining some key distinctions between minimal pairs of objects, we see that in the pairs {\it cube}/{\it rectangular prism} and {\it sphere}/{\it egg}, the latter object is lengthened along the X-axis.  Comparing {\it cylinder} and {\it capsule}, one has flat ends while the other has round ends.  Examining {\it cone} vs. {\it pyramid}, we see that even though the objects have the same dimensions, the faces of {\it pyramid} are all flat while everything but the base of {\it cone} is round.  These distinctions affect how each object behaves when it is interacted with.  The agent samples from the environment by stochastically placing the theme object on top of the cube and observing the result.

\begin{figure}[h!]
    \centering
    \includegraphics[trim=150 100 150 140, clip, width=1in]{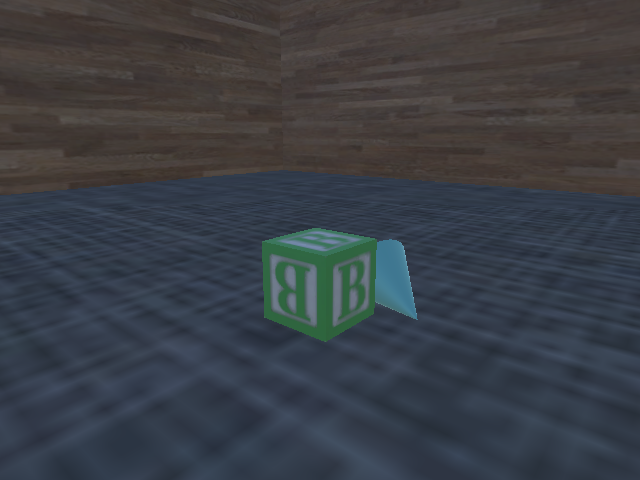}
    \includegraphics[trim=150 100 150 140, clip, width=1in]{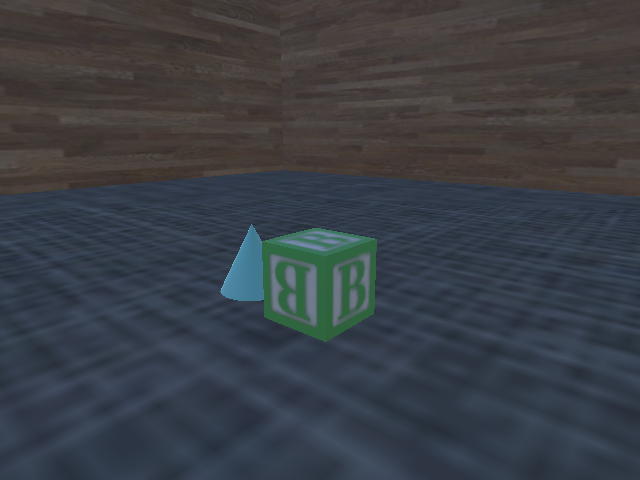}
    \includegraphics[trim=150 100 150 140, clip, width=1in]{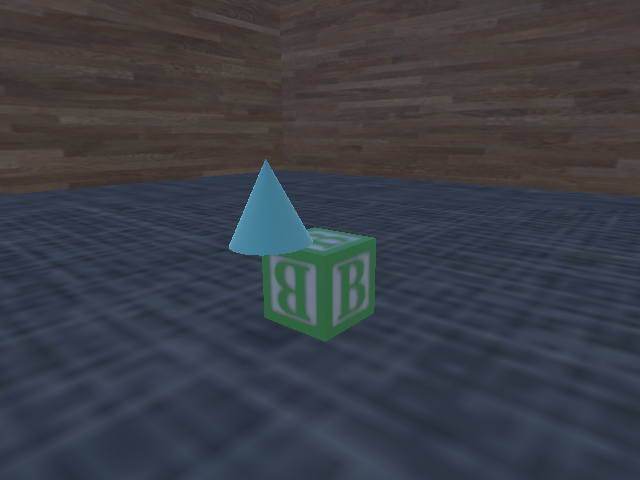}
    \includegraphics[trim=150 100 150 140, clip, width=1in]{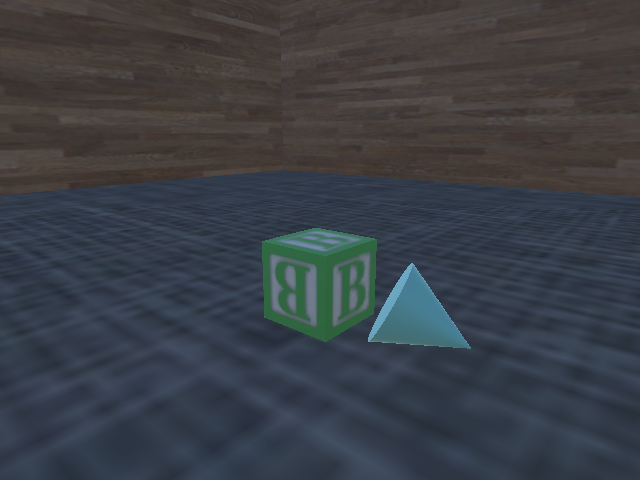}
    \includegraphics[trim=150 100 150 140, clip, width=1in]{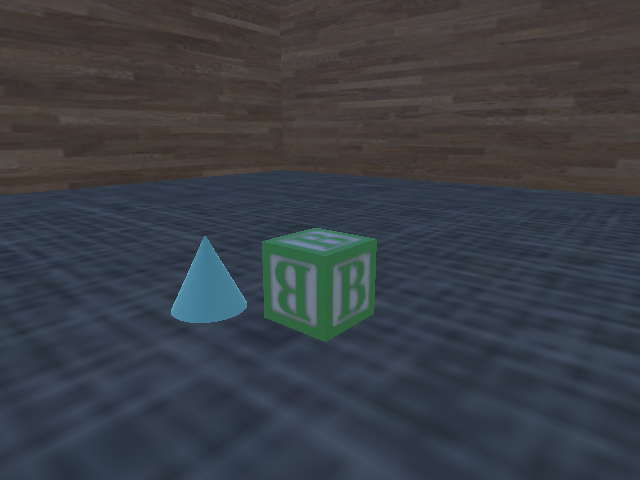}\\
    \includegraphics[trim=150 100 150 140, clip, width=1in]{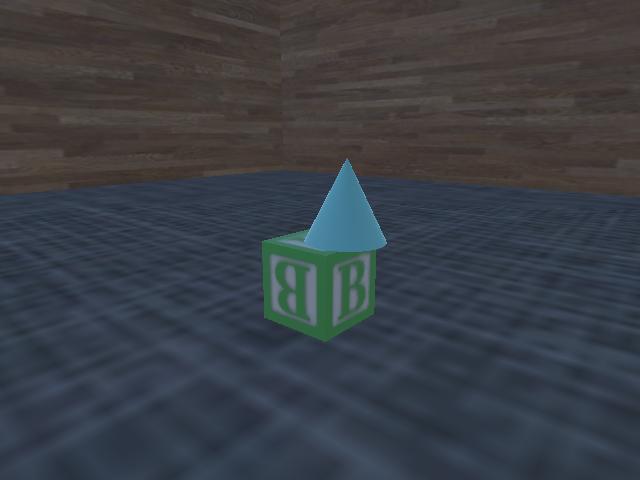}
    \includegraphics[trim=150 100 150 140, clip, width=1in]{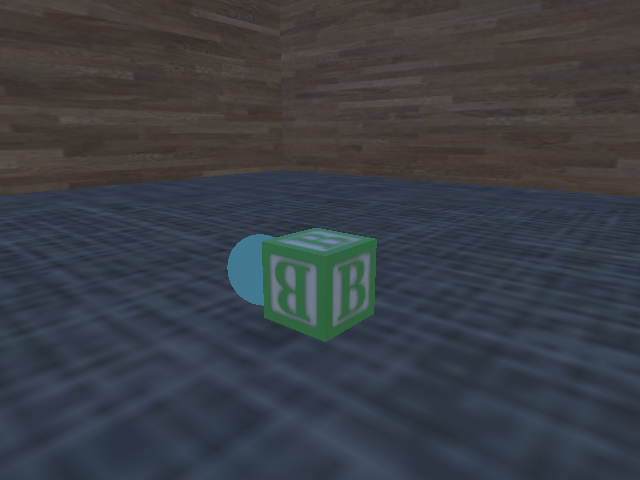}
    \includegraphics[trim=150 100 150 140, clip, width=1in]{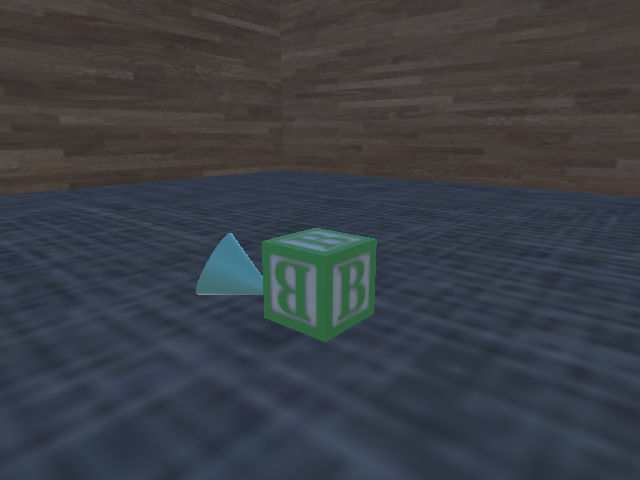}
    \includegraphics[trim=150 100 150 140, clip, width=1in]{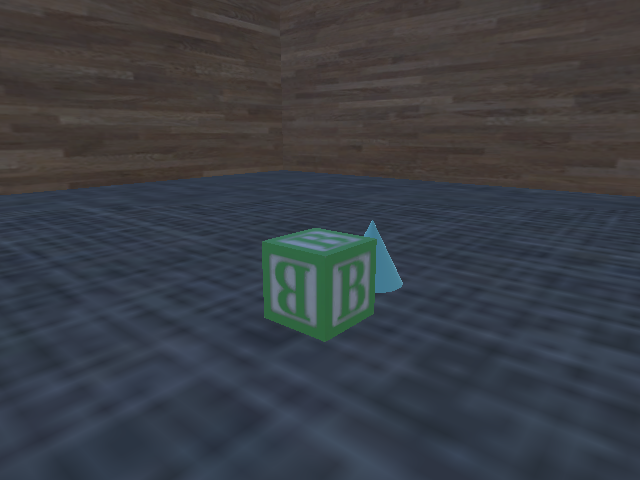}
    \includegraphics[trim=150 100 150 140, clip, width=1in]{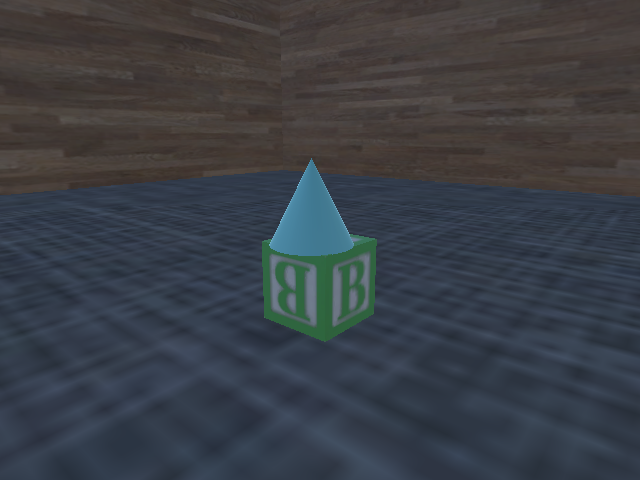}
    \vspace*{-3mm}
    \caption{10 cone stacking attempts.  When put in an unstable position, the cone falls off the cube.}
    \vspace*{-3mm}
    \label{fig:cone-10}
\end{figure}

If the object configuration that results from this placement is stable, the theme object will remain supported by the destination cube.  If not, it will fall off.  There are multiple reasons why an object might not stay balanced on top of the cube, which have to do with both placement location and the object properties themselves.  For instance, a cube will stay balanced if placed directly on top of the destination cube's surface, but if placed on the edge, will probably fall off.  A sphere may never stay balanced no matter where it is placed.  Meanwhile a cylinder or cone will probably stay balanced if appropriately placed {\it in an upright orientation}, but not in a horizontal one (e.g., see Fig.~\ref{fig:cone-10}).  These distinctions in configuration and resultant behavior correspond to the {\it habitats} \citep{pustejovsky2013dynamic} and {\it affordances} \citep{gibson1977theory} of the object, as they pertain to its ``stackability.''

\begin{wrapfigure}{l}{.4\textwidth}
\centering
\vspace{-2mm}
{\tiny
\def\baselinestretch{1.1}
$\avmplus{\att{cylinder}\\
	\attval{type}{\avmplus{
		\attvaltyp{head}{cylindroid}\\
		\attvaltyp{components}{nil}\\
		\attvaltyp{rotatSym}{$\{Y\}$}\\
		\attvaltyp{reflSym}{$\{XY,YZ\}$}
	}}                                
}$
\def\baselinestretch{1.9}
}
\vspace*{-2mm}
\caption{VoxML typing structure for a cylinder, showing axes and planes of rotational and reflectional symmetry.}
\label{fig:cylinder-typing}
\vspace*{-2mm}
\end{wrapfigure}

For each object type, we generate 10,000 samples of stacking attempts.  These attempts are not guided by a policy and no learning is taking place here (see Sec.~\ref{sec:novelty} for the integration of reinforcement learning).  Instead the agent stochastically places the theme object atop the destination object at a two dimensional coordinate in the range $(-1,1)$ that is scaled for the size of the destination object's surface and resolved to 3D coordinates in the VoxWorld environment.  VoxWorld's underlying Unity physics engine then simulates what happens to the objects in that configuration: either the theme object stays supported or it falls off.  The agent attempts to place the theme object on the destination object again, keeping it in the orientation it was after the previous attempt.  E.g., if an object placed in an upright orientation fell off the destination cube and landed on its side, the agent would attempt to place it again, still lying on its side.  After ten attempts, the theme object and destination object are re-placed randomly in the environment and sample gathering continues.

Because in a simulation, movements can be hyper-precise (unlike in the real world), even a clearly unstable object like a sphere can remain stably stacked due to a lack of noise or perturbations in the environment.  To overcome this and collect more realistic data, we apply a small ``jitter'' to the theme object once its placement is complete.  This jitter is to simulate the small force exerted upon an object when it is released by the grasper and may result in the object wobbling, sliding, or rolling, and potentially falling off the destination cube if it was not placed stably.  The direction of the jitter is drawn from VoxWorld's underlying VoxML semantics for objects \citep{pustejovsky2016voxml}.  For instance, if the theme object is a cylinder, its major axis is the Y-axis (Fig.~\ref{fig:cylinder-typing}) and so the post-action release jitter is applied perpendicular to the object's local Y-axis.  Therefore, this jitter implicitly encodes properties of the object habitats, such as orientations that do or do not facilitate rolling.  Objects like cubes or spheres, with no major axis, have the jitter applied in a random direction in the XZ plane.

After every placement attempt, the following numerical information is logged:
\begin{itemize}
    \vspace*{-2mm}
    \item Type of the theme object
    \vspace*{-2mm}
    \item Object rotation in radians at episode start
    \vspace*{-2mm}
    \item Radians between the world upright axis and the object upright (+Y) axis
    \vspace*{-2mm}
    \item Numerical action executed
    \vspace*{-2mm}
    \item Resulting spatial relations between the two objects
    \vspace*{-2mm}
    \item Object rotation and offset from world upright after the action (in radians)
    \vspace*{-2mm}
    \item Vector of the VoxML-derived post-action jitter
    \vspace*{-2mm}
    \item Position of the theme object relative to the destination object, before action, immediately after action, and after the object has had time to either settle in place, or roll off
    \vspace*{-2mm}
    
\end{itemize}

The resulting data numerically describes the behavior of objects during the stacking ``play''.  We also gather images of all attempts, mostly for visualization purposes (e.g., see Fig.~\ref{fig:cone-10}).  In this paper we focus on reasoning over the numerical data and how it allows us to build AI systems that can perform inference by ``embodying'' a position in the environment rather than processing images.

\vspace*{-4mm}
\section{Object Similarity Analysis}
\label{sec:obj-sim}
\vspace*{-2mm}

\begin{wrapfigure}{r}{.6\textwidth}
    \centering
    \vspace*{-4mm}
    \includegraphics[width=.6\textwidth]{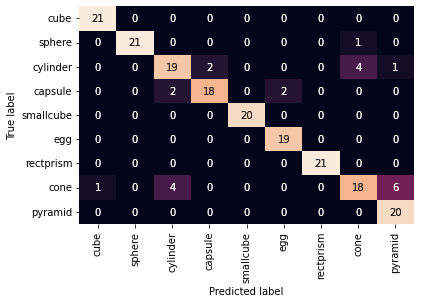}
    \vspace*{-6mm}
    \caption{Confusion matrix of behavior-based object classifier}
    \label{fig:ff-cm}
    \vspace*{-4mm}
\end{wrapfigure}

We first attempt to build a model to classify the objects based on their behavior in the freeform stacking play.  We use a 4-layer neural network (200, 100, 50, and 25 units respectively) with Leaky ReLU activation. We use weight decay of $0.01$. All the raw feature values listed above are the inputs, except for object type, which is the output to be predicted.  This model is trained on 14,400 total samples (1,600 per object type) using Adam optimization \citep{kingma2014adam} and cross-entropy loss with a learning rate of $0.0001$ and batch size of 32, for 200 epochs.  When tested on a further 3,600 samples (400 per object), the feedforward network achieves a test accuracy of {\bf 89.25\%}. Fig.~\ref{fig:ff-cm} shows the classifier's confusion matrix over 200 of the test samples.

\begin{figure}[h!]
    \centering
    \includegraphics[width=.48\textwidth]{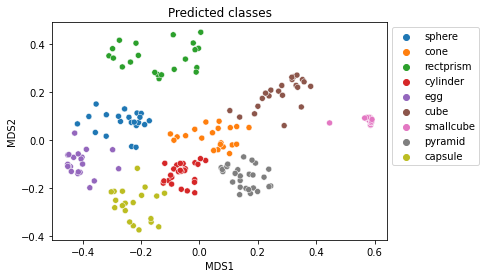}
    \includegraphics[width=.48\textwidth]{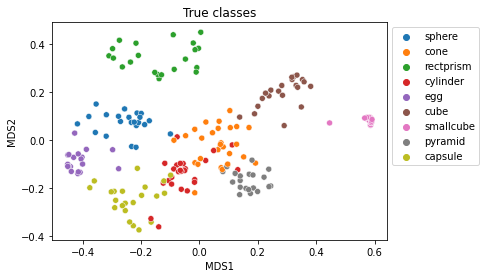}
    \vspace*{-3mm}
    \caption{MDS embeddings of last hidden layer activation, showing representational similarity. [L] colored by predicted label; [R] colored by true label.}
    \vspace*{-3mm}
    \label{fig:mds}
\end{figure}

To visualize how the network has clustered these object types, and for an intuitive sense of the relationships between the internal representations of the model, we retrieve the activation of the last hidden layer of the network and perform multidimensional scaling (MDS) \citep{borg2005modern} to visualize the similarity between individual final hidden layer representations.  Fig.~\ref{fig:mds} shows the MDS embeddings of the same aforementioned 200 test samples.  The left plot colors the points by the predicted label, and the right plot is colored by the true labels.

The MDS embeddings show that objects that share physical characteristics have similar representations.  For example, {\it sphere} and {\it egg}, the two most round objects in the object set, are neighbors.  The obvious neighbors to the {\it small cubes} are the larger cubes.  The confusions are also illustrative.  Figs.~\ref{fig:ff-cm} and \ref{fig:mds} show that one of more commonly confused objects is {\it cylinder}, it is relatively often confused with
    i) {\it capsule}, another cylindrical object, but with rounded ends; and
    ii) {\it cone}, another object with a single rounded face.
Figs.~\ref{fig:ff-cm} and \ref{fig:mds}[R] also show frequent confusion between {\it cone} and {\it pyramid}, because due to similar shapes, when they fall off the destination cube and come to rest ``sideways,'' they are in a similar orientation.  Therefore we can say the object behavior during the stochastic stacking attempts preserves information about each objects' physical characteristics, relevant configurations or habitats, and afforded behaviors (e.g., stackability or lack thereof).  In \cite{krishnaswamy2022exploiting}, we also explored object similarity through canonical correlation analysis (CCA), a linear approach, on the raw data. This paper further investigates the similarity of object types through MDS applied on the nonlinear hidden layer outputs of the neural classifier.

While the 9-way neural network classifier can solve the object classification problem well enough, these results do not tell us much about how the model is making these distinctions between objects based on their behavior.  What features are most important?  What concepts has it modeled to make the type distinctions?  In addition, most humans, particularly infants and toddlers, can ``train'' to accommodate a new type of object or item with far fewer examples, even if we allow that repeated encounters with the same item count as distinct stimuli.

\vspace*{-4mm}
\section{Transfer Learning to Accommodate New Classes}
\label{sec:tl}
\vspace*{-2mm}

To facilitate closer investigation of the object properties neural models learn from this embodied data, as well as to explore more sample-efficient models, we use a transfer learning approach. To begin, we train a deep feedforward architecture similar to that used in Sec.~\ref{sec:obj-sim} on 5,000 total samples of the classes {\it cube}, {\it sphere}, and {\it egg}.  The training features used here are the same as those used in Sec.~\ref{sec:obj-sim}. Between them, these three objects have three abstract properties that we hypothesize are important to distinguish between the 9 different objects in the entire dataset: flatness, roundness, and distinctions in axis of rotational symmetry. A model that is properly trained on these three objects should capture important features that can be used to distinguish additional objects as they are introduced.  This is analogous to an agent that has not encountered more than these three objects before but can use distinguishing features learned from its vocabulary of 3 objects to make inferences about other objects. Furthermore, this is analogous to an agent with the ability to do multiple tasks: one is identifying objects, the other is identifying concepts such as {\it round} and {\it flat}. We performed training on a Linux workstation with a 2.2 GHz Intel Xeon CPU (GPU training was not used). The number of training epochs is set to 100. We tune over 3 learning rates $\{0.001, 0.0001, 0.00001\}$. We early stop and choose the best learning rate using validation accuracy.

% but to prevent overfitting, we regularize the model with early stopping.

Subsequently, one object is added to the object vocabulary at a time.  To perform transfer learning from the source model to a model that accommodates one additional object, the first two hidden layers of the source model are frozen so additional training on new objects does not destroy important features learned from the original object set.  When transferring the knowledge from a source model to a target model, the target model can perform inference with few samples per object class. The success of this approach comes partly from the shared representation or structure in our data.

%We add one additional trainable layer, which introduces additional nonlinearities that might be necessary to distinguish the new object from the existing objects.

\begin{figure}[h!]
    \vspace*{-3mm}
    \centering
    % \subfloat[]{\includegraphics[height=.9in]{format/images/tl-3.png}}
    % \subfloat[]{\includegraphics[height=.9in]{format/images/tl_3_tmp.png}}
    \subfloat[]{\includegraphics[height=2in]{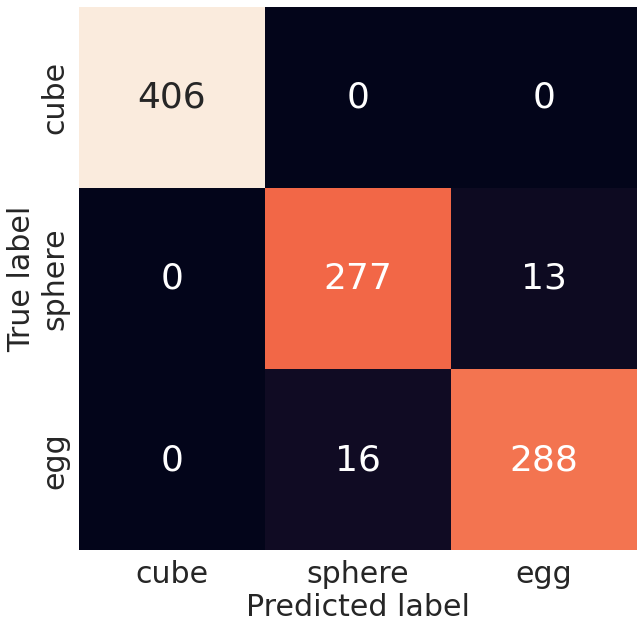}}
    % \subfloat[]{\includegraphics[height=.9in]{format/images/tl-4.png}}
    \subfloat[]{\includegraphics[height=2in]{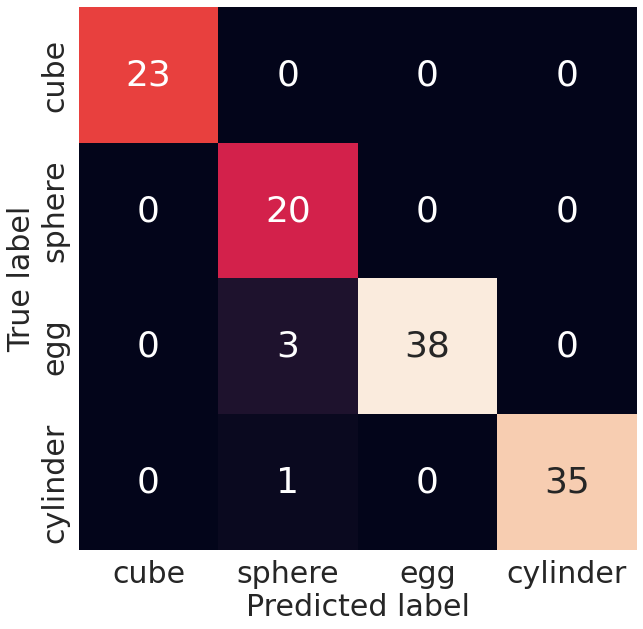}}
    % \subfloat[]{\includegraphics[height=.9in]{format/images/tl-5.png}}
    %\subfloat[]{\includegraphics[height=.9in]{format/images/output_3.png}}
    \subfloat[]{\includegraphics[height=2in]{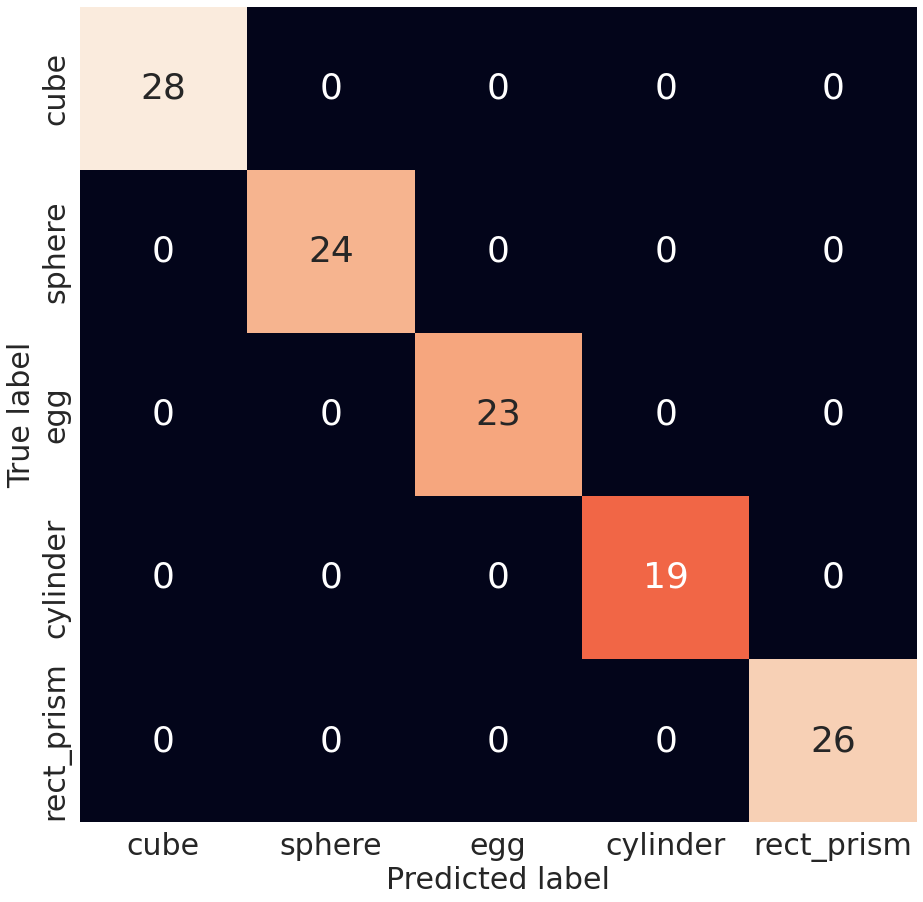}}\qquad
    % \subfloat[]{\includegraphics[height=.9in]{format/images/tl-6.png}}\qquad
    % \subfloat[]{\includegraphics[height=.9in]{format/images/output_4.png}}\qquad
    % \subfloat[]{\includegraphics[height=2.2in]{format/images/output_4_tmp.png}}
    \subfloat[]{\includegraphics[height=2.2in]{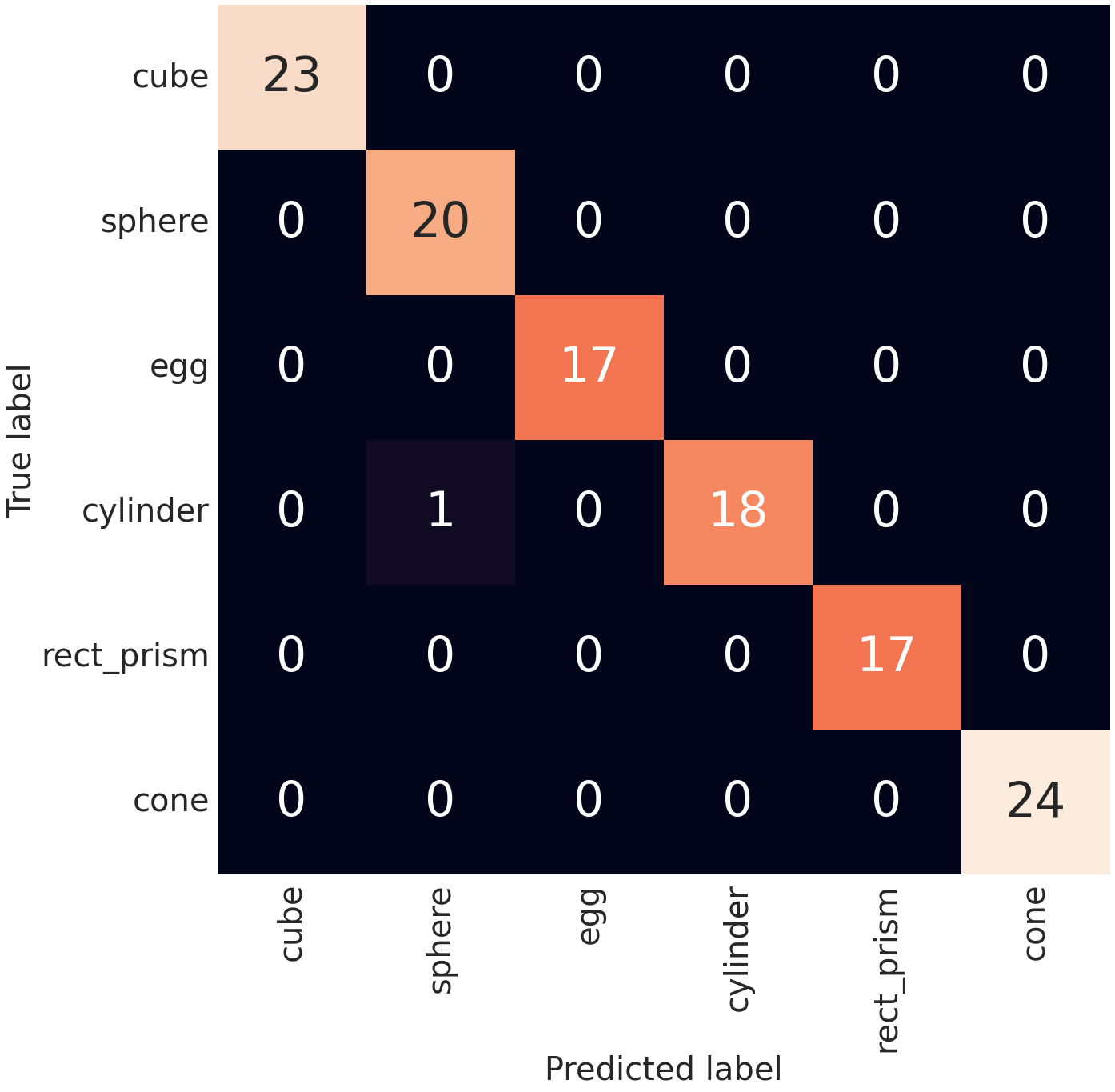}}
    % \subfloat[]{\includegraphics[height=.9in]{format/images/output_5.png}}
    % \subfloat[]{\includegraphics[height=2.2in]{format/images/output_5_tmp.png}}\qquad
    \subfloat[]{\includegraphics[height=2.2in]{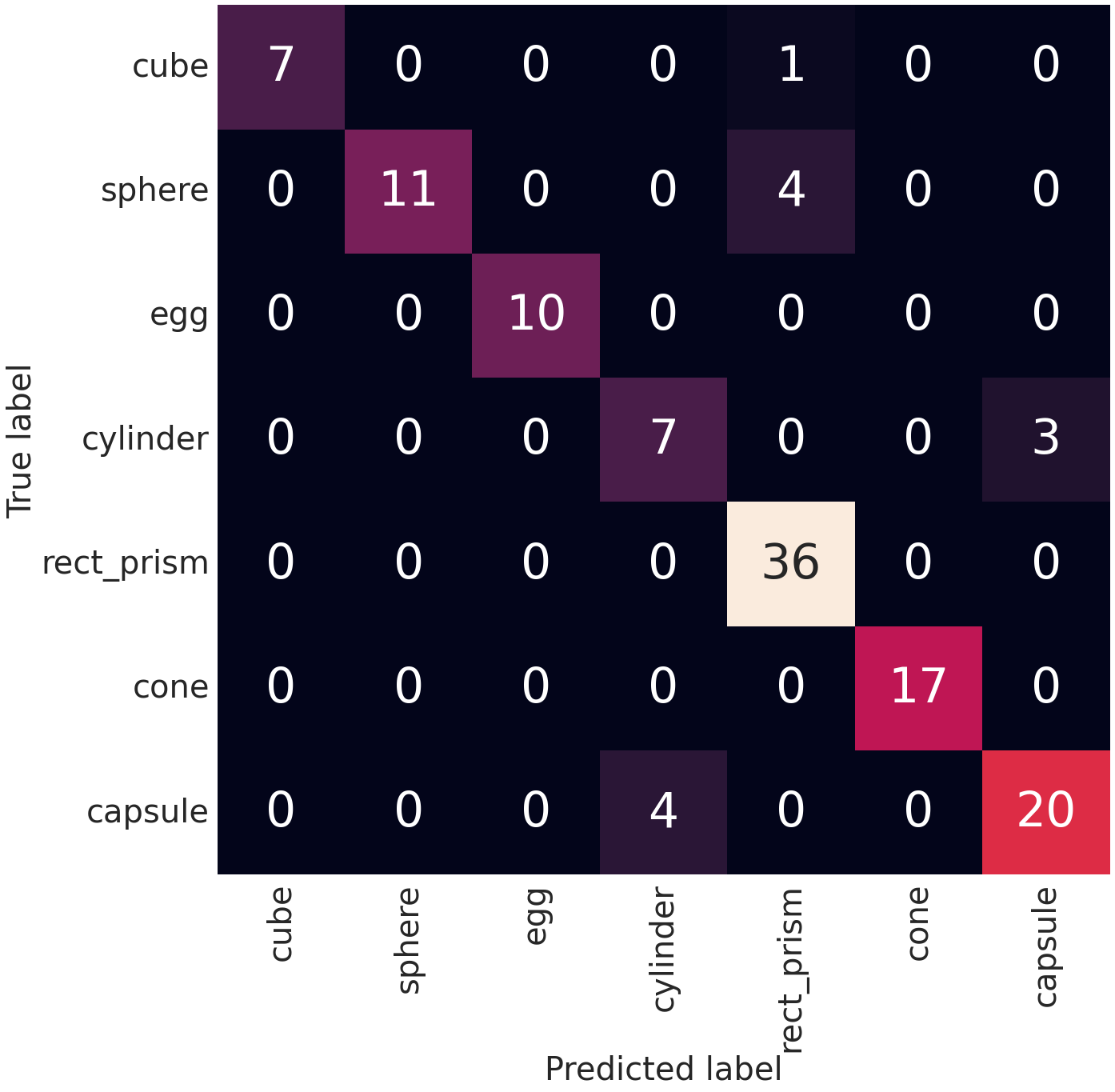}}\qquad
    % \subfloat[]{\includegraphics[height=2.2in]{format/images/output_6.png}}
    % \subfloat[]{\includegraphics[height=2.2in]{format/images/8cls.png}}
    \subfloat[]{\includegraphics[height=2.2in]{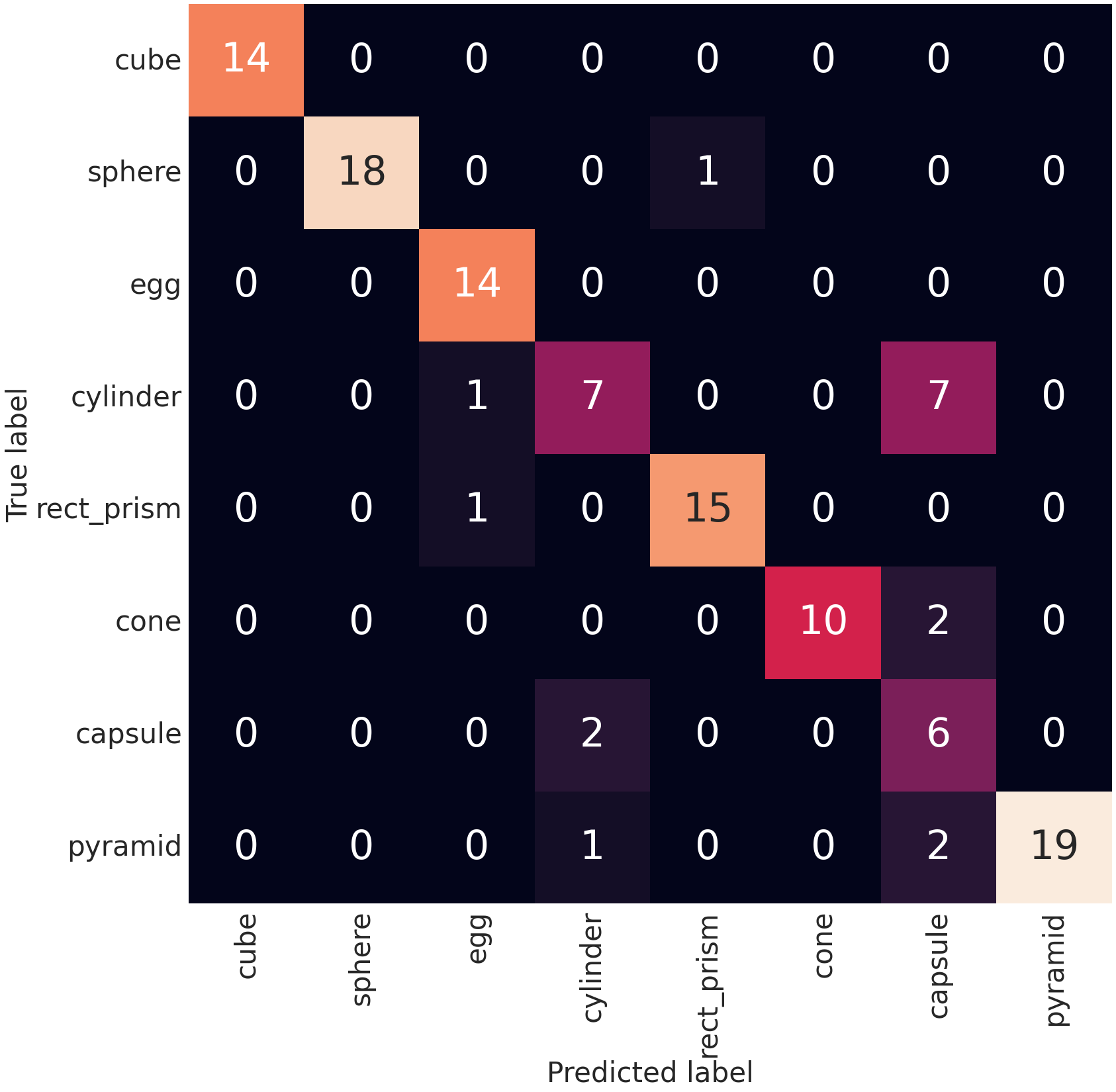}}
    % \subfloat[]{\includegraphics[height=.9in]{format/images/output_7.png}}
    % \subfloat[]{\includegraphics[height=2.2in]{format/images/output_7_tmp.png}}
    % \subfloat[]{\includegraphics[height=2.2in]{format/images/9cls.png}}
    \subfloat[]{\includegraphics[height=2.2in]{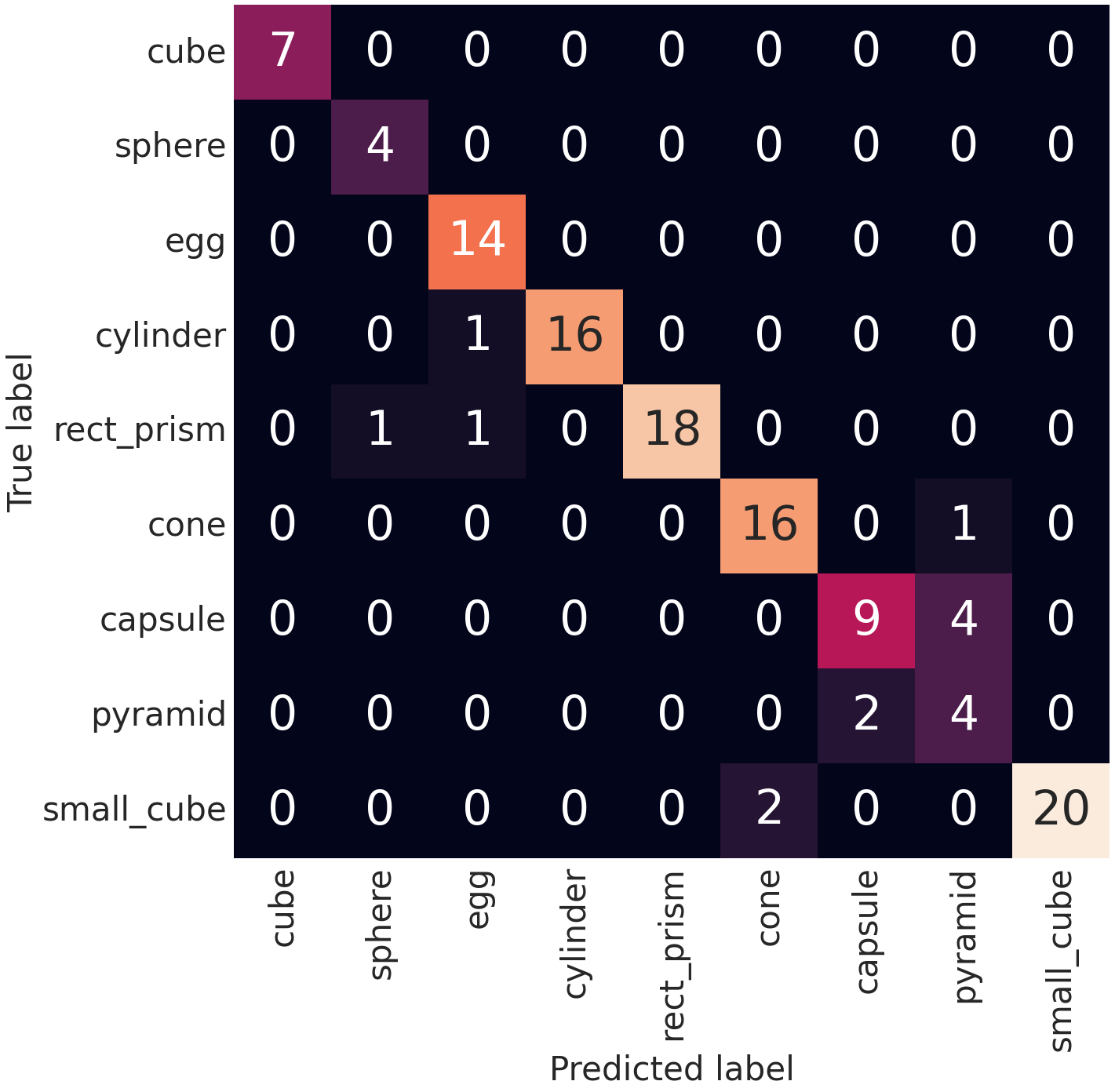}}
    %\subfloat[]{\includegraphics[height=1.25in]{format/images/tl-7.png}}
    % \subfloat[]{\includegraphics[height=1.25in]{format/images/tl-8.png}}
    % \subfloat[]{\includegraphics[height=1.25in]{format/images/tl-9.png}}
    \vspace*{-3mm}
    \caption{Confusion matrices of transfer-learned models with progressively more object types.  (a) base, (b) {\it +cylinder}, (c) {\it +rect. prism}, (d) {\it +cone}, (e) {\it +capsule}, (f) {\it +pyramid}, (g) {\it +small cube}.}
    \vspace*{-3mm}
    \label{fig:tl-progression}
\end{figure}

In the target model, we add a hidden layer on top of the source model's last hidden layer. As the first two layers are frozen, training this hidden layer better facilitates training convergence when using a low number of samples. The information captured in the source model can be successfully transferred to a target model regardless of number of classes in the target model.  We modify the softmax layer and fine tune the source model, trained on $k-1$ objects, to the target model for $k$ objects.  With each additional object, the previous target model becomes the source model and this process is repeated. Fig.~\ref{fig:tl-progression} shows the evaluation results.

%At the very first step a model has been trained on the 5000 samples in total including egg, sphere and cube. The model consists of 4 hidden layers (Note to Nikhil: this is better not to be in the draft. I think reviewers won't be nice in general because we are using TL without a  large vision/language model because of our data. Second we have another model with more layers but the architecture is not used for our MDS plots). 
%Layers prior to 3rd hidden layer are frozen to avoid destroying the general information/feature they contain during future training rounds. One new trainable layer is added on top of the frozen layers every time on a new data ( in our context the added object). The new layers help to turn the old features into predictions on a new dataset. When transferring features from source model to the target model, we use 600 samples in total. But, due to shared structure in our data, we can reduce the number of samples to 30 (which in turn leads to the lower amount of samples per class) in the target model and good model performance would be still maintainable

\begin{wrapfigure}{l}{.5\textwidth}
    \centering
    \vspace*{-6mm}
    \includegraphics[width=.5\textwidth]{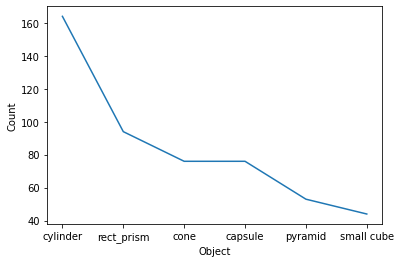}
    \vspace*{-6mm}
    \caption{fine tuning samples per object.}
    \label{fig:samples-vs-obj}
    \vspace*{-3mm}
\end{wrapfigure}

We are able to maintain high classification accuracy by incrementally introducing one novel object and fine tuning the source model for that objects, arriving at a final test accuracy of {\bf 90\%} when all 9 object types are incorporated.  Each time we test on a different randomly-selected subset of object samples.  When transferring features from source model to target model, we use a constant 600 samples in total.  Interestingly, this means that as new objects are added, the number of samples {\it per object} used for fine tuning goes down, as shown in Fig.~\ref{fig:samples-vs-obj}, suggesting that the base model sufficiently captures important discriminative features, and that only a few samples of new objects are needed for our model to progressively accommodate them.  We hypothesize that because the source task starts with objects that collectively have both flat and round features, the ability to distinguish objects based on these features is captured in the initial source task and can be shared in the following target tasks. Therefore we hypothesize that the order of object addition is unlikely to matter much.

Interestingly, the transfer-learned model (without the small cube) shows similar confusions to the initial feedforward model trained on all nine objects (see Sec.~\ref{sec:obj-sim}).  When the small cube is added, these confusions are lessened significantly (compare Figs.~\ref{fig:tl-progression}(f) and \ref{fig:tl-progression}(g) to Fig.~\ref{fig:ff-cm}).  The small cube introduces a height contrast with the larger cube from the base model and, since all other objects are of a similar size to the larger cube, the height of the small cube contrasts with the height of all other objects.  Height is not directly recorded in the dataset, but is implicitly encoded in the position of the object, as the Y-coordinate of the small cube is lower than the Y-coordinate of any other object in the same conditions.

\begin{wrapfigure}{r}{.6\textwidth}
    \centering
    \vspace*{-3mm}
    \includegraphics[width=.6\textwidth]{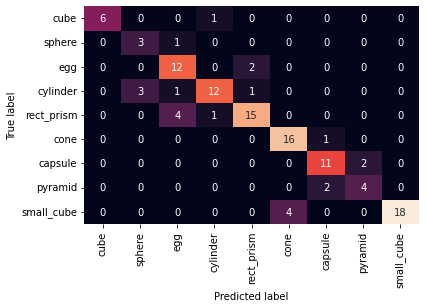}
    \vspace*{-6mm}
    \caption{Confusion matrix for 10-layer model of constant size after transfer learning is performed up to 9 object types (cf. Fig.~\ref{fig:tl-progression}(g)).}
    \label{fig:tl-largemodel-9}
    \vspace*{-5mm}
\end{wrapfigure}

By the end of the transfer learning process, we have a model with 10 hidden layers that has progressively grown with the incorporation of each new object type.  We therefore compare the performance of this model to a 10-layer model whose size remains constant through all transfer learning procedures.  We repeat the same transfer learning procedure as above with this model architecture, without adding new layers through the process. Here, the only change is to add a node to the softmax layer.

Fig.~\ref{fig:tl-largemodel-9} shows the results from this model, with a test accuracy of {\bf 80.83\%}.  This static model makes significantly more mistakes than the dynamically growing model, perhaps due to overfitting to cube, sphere, and egg-related features when training a larger base model.

\vspace*{-4mm}
\section{Inferring Abstract Concepts}
\label{sec:abstract}
\vspace*{-2mm}

The success of transfer learning in this task demonstrates the effectiveness of the VoxWorld embodied simulation environment for gathering data to enable the metacognitive task of model expansion through transfer learning.

The objects in the dataset, however, are not just instances of multiple different classes, but also display properties and contrasts that inhere across multiple object classes.  As discussed earlier, we have both round objects and flat objects, as well as objects like cylinder and cone, that display both roundness and flatness depending on their orientation. In this section, we test the capacity of the previously-learned model to make inferences about these more abstract properties.

In the gathered data, we store the rotation of objects after the placement action is taken.  Due to the geometric properties of the object, the rotation is a deterministic indicator of how the object comes to rest.  For example, a cylinder resting at either 0 or $\pi$ radians is upright, on its flat end, while being at either $\frac{\pi}{2}$ or $\frac{3\pi}{2}$ radians means it is horizontal, or resting on its round edge.  For a cone, resting at a rotation of 0 radians means it is resting on its flat base, while resting at $\sim\frac{2\pi}{3}$ radians means it is resting on its round side (e.g., see Fig.~\ref{fig:cone-10}).  All other objects are either entirely flat-faced or have only round sides.  This means that each individual sample of a stacking attempt can be classified into the object coming to rest on a {\it flat} side or a {\it round} side.  If the previously-trained object classification model was successful in part due to learning features correlated with roundness or flatness, then it should also be useful for this round side/flat side classification task.

\begin{wrapfigure}{l}{.4\textwidth}
    \centering
    \vspace*{-3mm}
    \includegraphics[width=.4\textwidth]{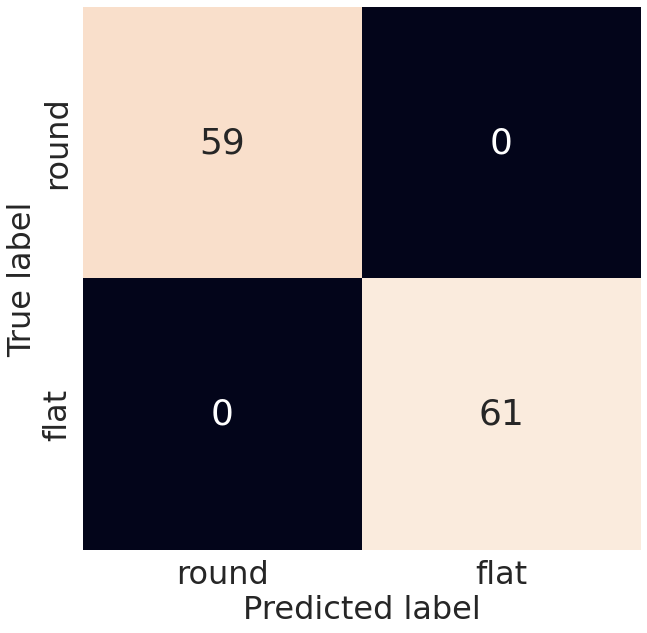}
    \caption{Confusion matrix for round side vs. flat side classification.}
    \vspace*{-3mm}
    \label{fig:tl-round-flat}
    % \label{fig:round_flat}
\end{wrapfigure}

The fine-tuning procedure here reflects the previous procedure for fine tuning for new object classes.  We add one additional hidden layer on top of the last hidden layer of the now 10-layer source model (from Fig.~\ref{fig:tl-progression}) to allow for additional nonlinearities between the source model's object-level features and abstract flatness/roundness features, and we fine tune the resulting target model on 300 flat-edge object samples and 300 round-edge object samples, distributed across multiple object types.  Fig.~\ref{fig:tl-round-flat} shows a test accuracy of {\bf 100\%} on a further 120 test samples, demonstrating that the features learned during object classification can also be used to discriminate more abstract categories like ``roundness'' and ``flatness.''

\iffalse
\begin{wrapfigure}{r}{.6\textwidth}
    \vspace*{-3mm}
    \centering
    \begin{tabular}{|l|l|l|l|}
        \hline
        {\bf Task} & {\bf \# classes} & {\bf Model} & {\bf Test acc.} \\
        \hline
        Obj. class & 9 & Baseline & 89.25\% \\
        Obj. class & 9 & Static Transfer & 80.83\% \\
        Obj. class & 9 & Dynamic Transfer & 92.50\% \\
        Flat/round & 2 & Dynamic Transfer & 98.33\% \\
        \hline
    \end{tabular}
    \vspace*{-3mm}
    \caption{Summary of transfer learning tasks and models.}
    \vspace*{-3mm}
    \label{tab:transfer-summary}
\end{wrapfigure}

Table~\ref{tab:transfer-summary} shows a summary of the transfer learning tasks, models, and performances discussed in the previous sections.  ``Baseline'' refers to the model trained from scratch directly on 9 objects.  ``Static Transfer'' refers to the transfer learning model that remains a constant size without additional trainable hidden layers.  ``Dynamic Transfer'' refers to the model that has trainable hidden layers added to it to with additional object classes.
\fi

\vspace*{-4mm}
\section{Detecting Novel Classes}
\label{sec:novelty}
\vspace*{-2mm}

Previously, we discussed how transfer learning can be used to make a model accommodate new object classes based on their behavior.  However, there we always explicitly incremented the number of objects with each transfer learning iteration.  In this section, we discuss how a model can automatically detect when a new object class has been introduced into the environment, so that an agent can trigger its own transfer learning procedure to accommodate the new object type.  This is a metacognitive procedure where the system detects when its own model is inadequate to the environment, so it can use methods like those discussed in Sec.~\ref{sec:tl} to adapt to new conditions.

Our methodology involves 1) training a policy to perform a task with a known object type; 2) attempting to use any object presented in the same task using the aforementioned trained policy; 3) observing differences in behaviors of the various objects and using those differences to identify if an instance of an object is sufficiently different from known objects to likely constitute a new class.

At slightly more than 6 months old, most infants appear able to intuit that an object will not fall if supported from the bottom on over 50\% of its lower surface \citep{baillargeon1992development,dan2000development,huettel2000effects,spelke2007core}. We use VoxWorld's integration with Unity ML-Agents \citep{juliani2018unity}, OpenAI Gym \citep{brockman2016openai}, and the Stable-Baselines3 reinforcement learning (RL) package \citep{raffin2021stable} to train an RL algorithm that resembles this intuition in the block stacking task.  Learning to stack is, of course, not a novel task in the RL community (cf. \cite{lerer2016learning}, \cite{li2017acquiring}, \cite{li2020towards}, \cite{hundt2020good}, just to name a few).  While it is a useful task for demonstrating RL algorithms and AI's capability to learn representations of certain physical intuitions, the work we present here also demonstrates how an RL model for this relatively simple task, coupled with embodied simulation, can be used to drive computational implementations of certain metacognitive processes.  All training in this experiment was performed using a MacBook Pro laptop with an Intel CPU.

\vspace*{-3mm}
\subsection{Policy Training}
\vspace*{-2mm}

\begin{wrapfigure}{r}{.4\textwidth}
    \vspace*{-3mm}
    \centering
    \includegraphics[height=.9in]{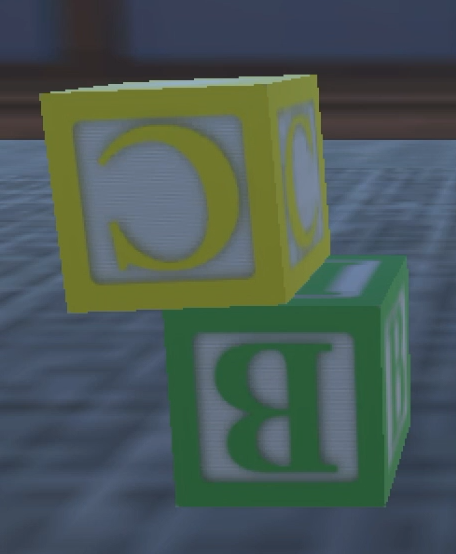}
    \includegraphics[height=.9in]{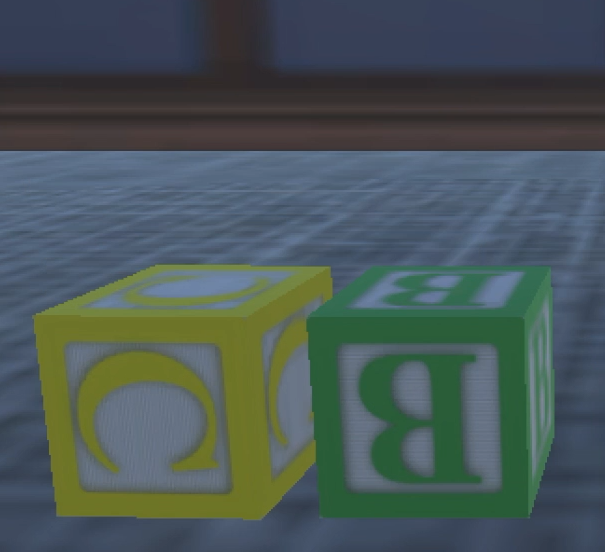}
    \includegraphics[height=.9in]{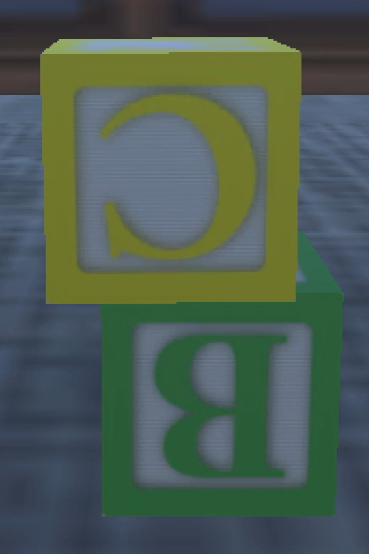}
    \vspace*{-6mm}
    \caption{Unsuccessful (left, center), and successful (right) stacking.}
    \vspace*{-3mm}
    \label{fig:stack-training}
\end{wrapfigure}

We first train a TD3 policy \citep{fujimoto2018addressing} to stack two equally-sized cubes.  One cube is the destination object and the other is the theme object.  The scaled action space is a 2D continuous space $[0,1000]\times[0,1000]$.  By default the optimal action is $(500,500)$ (though this can be perturbed in VoxWorld to test generalization).  The agent's goal is to place the theme block on the destination such that the two-block stack stays up (Fig.~\ref{fig:stack-training}). 

\iffalse
After the action is complete, we also add a small ``jitter'' to the object.  Since our agent simply moves blocks in space in the virtual environment, this simulates the small force that a real embodied agent (i.e., a toddler, or a robot) would exert upon the object when releasing it. This makes the simulation more realistic. This jitter force is applied perpendicular to the major rotational axis of the theme object if one exists.  Since cubes are symmetrical along all 3D axes, this force in training is applied in a random direction.
\fi

The state space comprises the number of blocks in the stack (an integer 1 or 2), and the 2D center of gravity of the stack (X and Z values only) relative to the center of the destination object. The agent receives a reward of -1 for missing the destination block entirely, of 9 for touching the surface of the destination block but not stacking stably, and up to 1,000 for stacking the two blocks successfully (with a discount of 100 for each additional attempt---i.e., 900 on the second try, etc.).

\vspace*{-3mm}
\subsection{Policy Evaluation and Data Gathering}
\vspace*{-2mm}

\begin{wrapfigure}{l}{.6\textwidth}
    \centering
    \includegraphics[width=.6\textwidth]{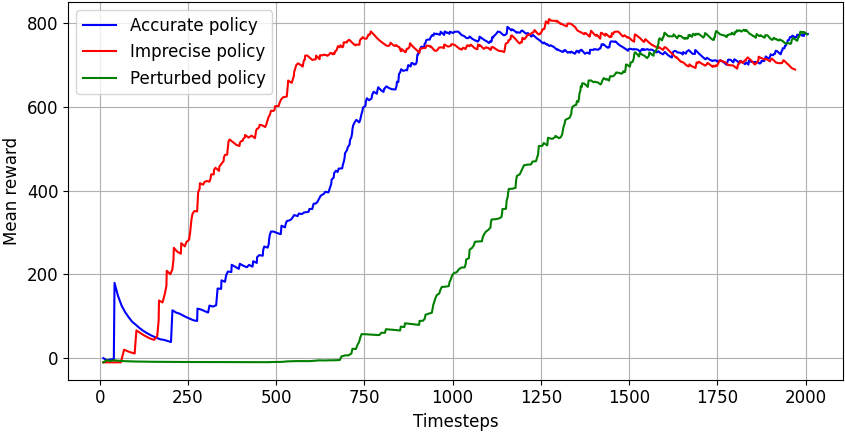}
    \vspace*{-3mm}
    \caption{Mean reward vs. timesteps during training.}
    \vspace*{-3mm}
    \label{fig:rl-training}
\end{wrapfigure}

Fig.~\ref{fig:rl-training} shows reward plots for policies trained using this method.  Training the RL policy completes in under 30 minutes. The two policies that we evaluate when gathering data for this task are represented by the left two curves: the {\bf accurate policy}, where the trained policy is very close to the optimal action, and the {\bf imprecise policy} where the trained policy is slightly less well-optimized and the theme cube falls off somewhat more often in testing.  In the plot where the reward starts climbing around timestep 700, the action space was perturbed so the optimal policy is far from the center.  We do not focus on this policy since downstream results are similar to the others.

We then evaluate the trained policies using sets of different theme objects and gather data about each evaluation.  Since the policies were trained to only stack a cube on another cube, this is tantamount to making the agent attempt to stack various objects {\it as if they are all cubes}.

\begin{figure}[h!]
    \centering
    \includegraphics[width=.48\textwidth]{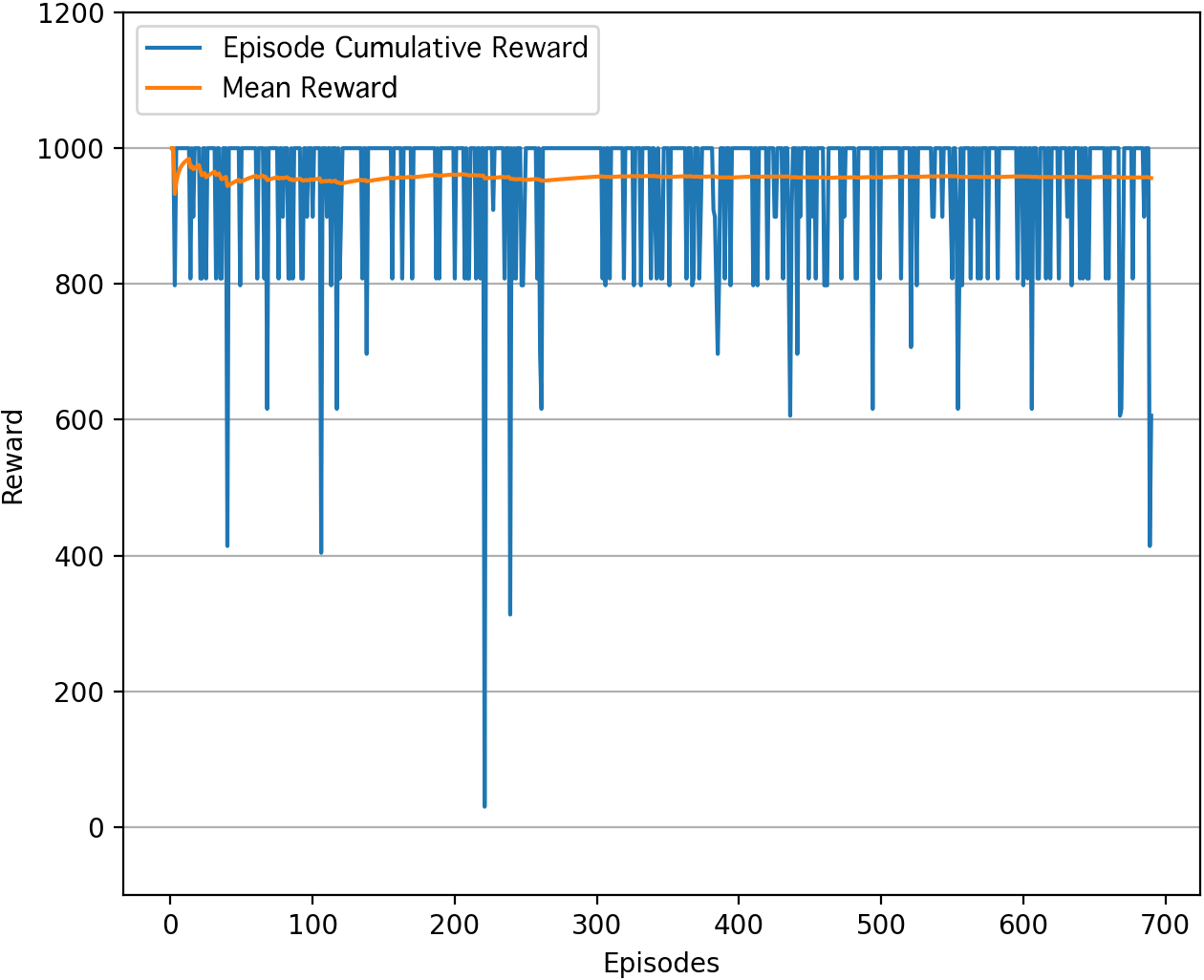}
    \includegraphics[width=.48\textwidth]{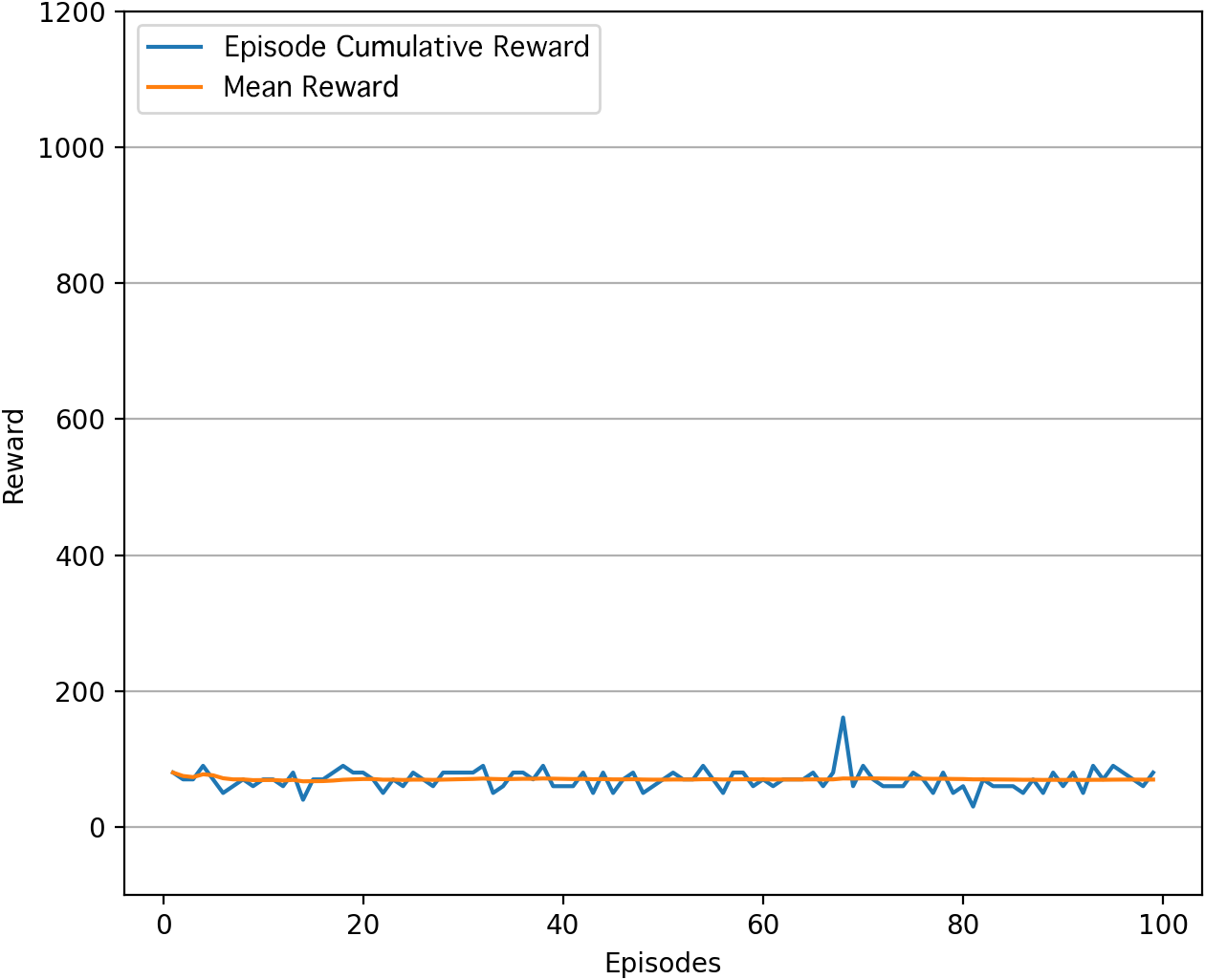}
    \includegraphics[width=.48\textwidth]{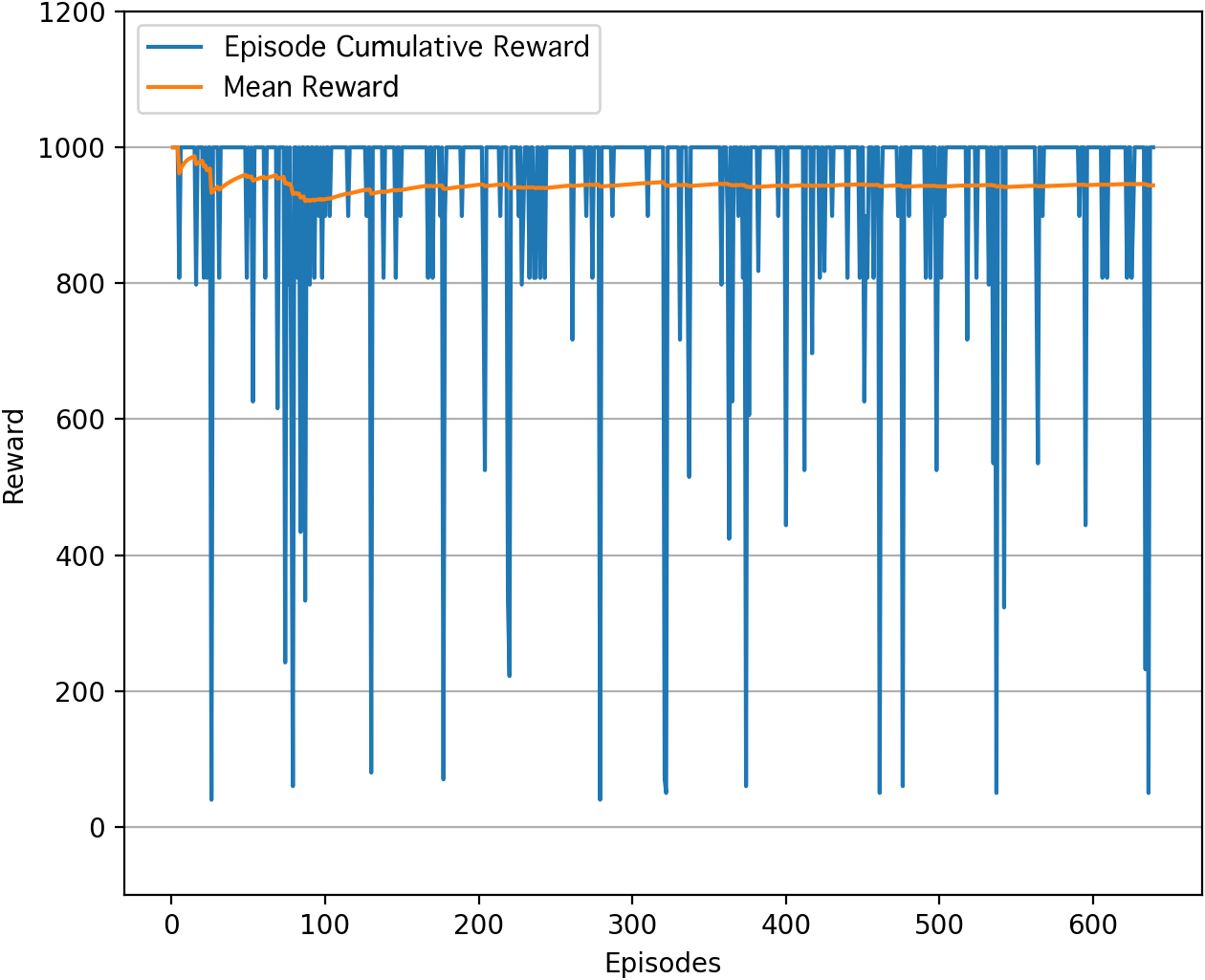}
    \includegraphics[width=.48\textwidth]{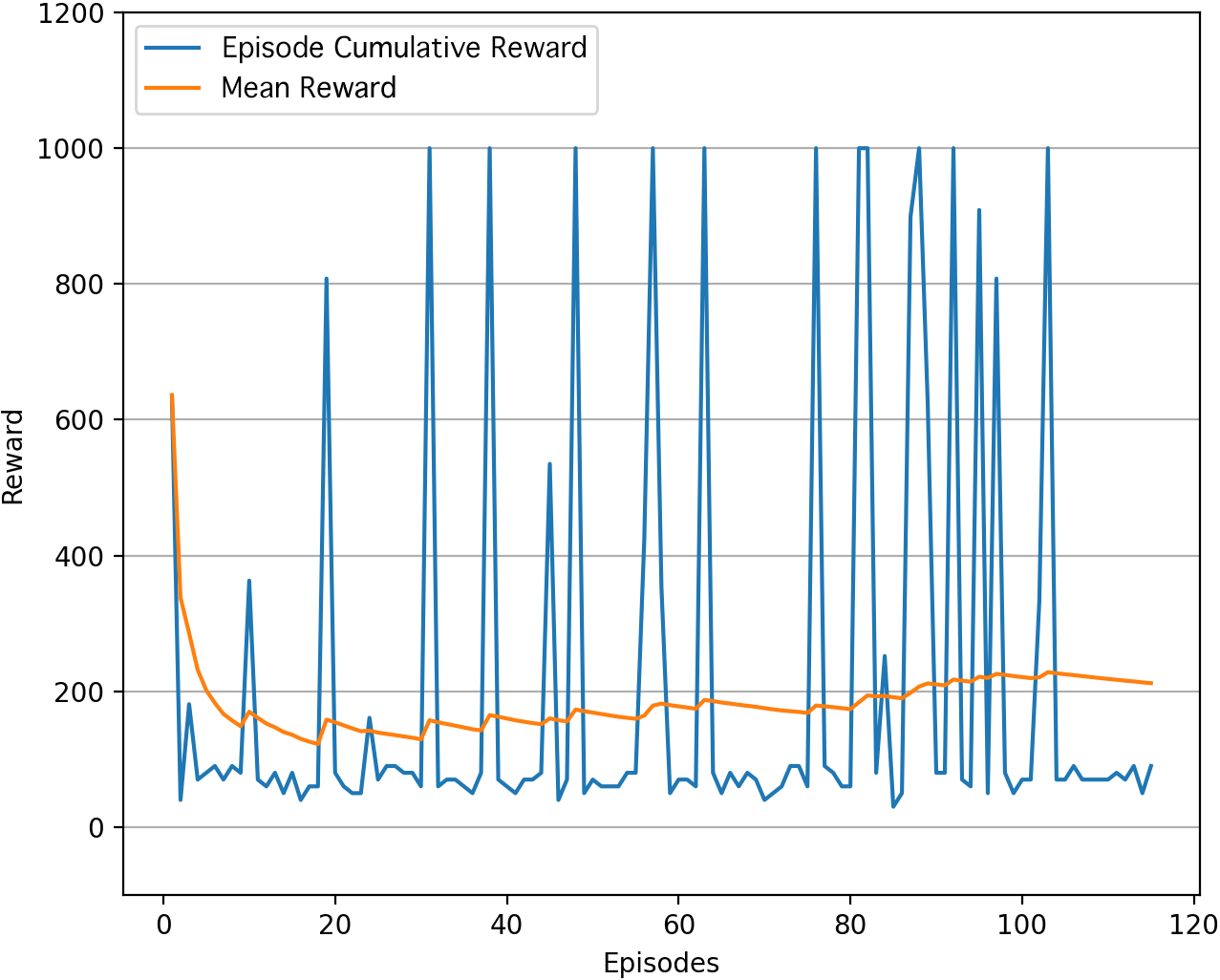}
    \vspace*{-3mm}
    \caption{Accurate policy reward plots for stacking (from top left) cube, sphere, cylinder, capsule during evaluation. Episodes are on the X-axis and rewards are on the Y-axis. Plot for small cube is very similar to cube.}
    \vspace*{-3mm}
    \label{fig:reward-plots}
\end{figure}

We evaluate each policy for 1,000 timesteps with the objects {\it cube}, {\it sphere}, {\it cylinder}, {\it capsule}, and {\it small cube}.  We focus on 5 objects here to examine how the behavior of novel objects can be exposed given a minimal set of objects that preserve distinctions in the ``roundness'' and ``flatness'' properties. Fig.~\ref{fig:reward-plots} shows the evaluation reward plots for the accurate policy (the blue line is the reward after each episode and the orange line is the mean cumulative reward), and show that the cube is the easiest object to stack, followed by cylinder, capsule, and finally sphere. As during training, the agent gets 10 attempts to stack per episode, so stackable objects like cube and cylinder can complete more episodes in 1,000 evaluation timesteps.

During policy evaluation, we gather the same information about each stacking attempt as mentioned in Sec.~\ref{sec:environment-data}, as well as the reward for the attempt, the cumulative total reward over the episode, the cumulative mean reward over the episode, and the height of the stack (1 or 2 objects) after each attempt. We gathered two datasets, one each using the accurate and imprecise policy.

\vspace*{-3mm}
\subsection{Novel Class Detection}
\vspace*{-2mm}

We want to be able to give an algorithm a model of a subset of these classes (e.g., cube and sphere) and have it identify that a new type of object (e.g., cylinder or capsule) is different from any of the known classes based on the way it behaves when interacted with (i.e., stacked).  We also want to be able to identify that new samples of a known class (e.g., small cubes), are not new classes of objects, but additional instances of a known class.  Here we do not consider size as a distinguishing feature in the model, only object behavior when stacked.

To detect a novel class, we 1) identify which known class an object is most similar to; and 2) determine if the new object is different enough from the most similar known class to be considered likely novel.  We start by training a classifier on two known classes: cube and sphere, two of the most dissimilar objects according to similarity analysis in Sec.~\ref{sec:obj-sim} or CCA \citep{krishnaswamy2022exploiting}.  We use a 1D convolutional neural net.  We group inputs by episodes and to maintain a balanced sample, use only the first 90 evaluation episodes, reserving a further 10 as a development set for testing the classifier, and the remainder of the data for detecting novel classes.  Since episode length can vary based on the number of attempts, we pad out the length of each input to 10 timesteps, copying the last sample out to the padding length.  Therefore an episode where the policy stacked the object successfully on the first try will consist of 10 identical timestep representations, while an episode where the agent tried and failed the stack the object 10 times will have 10 different timestep representations.

\iffalse
\begin{wrapfigure}{l}{.4\textwidth}
    \centering
    \includegraphics[width=.4\textwidth]{images/cnn-diagram.png}
    \vspace*{-3mm}
    \caption{1D CNN object classifier.}
    \vspace*{-3mm}
    \label{fig:cnn}
\end{wrapfigure}
\fi

The classifier consists of two convolutional layers (256 and 128 units). The filter size in the first layer is $c$, a variable equal to the number of parameters saved at each evaluation timestep during data gathering ($c = 19$ here) and a stride length of 8, and the second layer uses a filter size of 4 and a stride length of 2.  This allows the convolutions to generate feature maps in the hidden layers that are approximately equal to the size of a single timestep sample, and convolving over this approximates observing each timestep of the episode in turn.  The convolutional layers are followed by two 64-unit fully-connected layers and a softmax layer.  All hidden layers use ReLU activation. We train for 500 epochs using Adam optimization, a batch size of 100 (10 episodes) and a learning rate of $0.001$.

%Since the differences between cubes and sphere in their stacking behavior are so evident, this simple two-class classifier can routinely achieve 100\% evaluation accuracy on the dev-test set.

We then take batches drawn from classes unknown to the model, e.g., cylinder and capsule, and from additional instances of known classes, e.g., cubes, spheres, and small cubes.  The classifier, trained over only two classes, will classify even the non-sphere or cube samples as sphere or cube.  Most commonly, cylinder is classified as cube and capsule is classified as sphere, because these objects' stacking behavior is similar.  Small cube is (correctly) classified as a cube.

We then retrieve the 64-dimensional embedding vectors for each sample in the testing batch, and for each sample of the most similar known class.  The embedding vectors of the new samples can be compared to embedding vectors of known instances of that predicted class to determine if this new batch is similar enough to truly be the same as the known class or not.

We compute $\vec{\mu_S}$ and $\vec{\sigma_S}$, the mean embedding vector of the known class and the standard deviation of the known class vectors, respectively, as well as $\vec{\mu_N}$, the mean embedding vector of the new batch. Then, assuming that if all samples, new or known, were in fact members of the same class, there would still be some outliers, we find individual outliers in the new batch samples and in the known class samples by dividing the cosine distance between $\vec{\mu_S}$ and the sample $\vec{v}$ in question by the cosine distance between $\vec{\mu_S}$ and $\vec{\mu_S}+\vec{\sigma_S}$. Let $\rho_{\vec{v}} = \frac{\text{cos}(\vec{\mu_S},\vec{v})}{\text{cos}(\vec{\mu_S},\vec{\mu_S}+\vec{\sigma_S})}$, and if $\rho_{\vec{v}} > 1$, the sample $\vec{v}$ is considered to be an outlier $\vec{o} \in O$, where $\rho_{\vec{o}} = \frac{\text{cos}(\vec{\mu_S},\vec{o})}{\text{cos}(\vec{\mu_S},\vec{\mu_S}+\vec{\sigma_S})}$.  Let $O_S$ be the set of outliers among the samples of the known class $S$ and let $O_N$ be the set of vectors in the new batch $N$, where $\rho_{\vec{o_N}} > 1$.  Outlying samples may still belong to the known class (e.g., sometimes a cube simply fails to stack properly due to bad placement, not its properties, but nonetheless appears to be very different from other cubes in terms of its behavior), so we perform Z-score filtering on the computed outliers, using a Z threshold of 3, and $\mu_\rho$ and $\sigma_\rho$, the mean and standard deviation, respectively, of the previous computations over the outlier vectors.  If $\frac{(\rho_{\vec{o}}-\mu_\rho)}{\sigma_\rho} \geq 3$, $\vec{o}$ is removed from the set of outlier embeddings.  For all outliers $\vec{o_N} \in O_N$ that were derived from new batch samples and all outliers $\vec{o_S} \in O_S$ derived from known class samples, we sum $\rho_{\vec{o_N}}$ for all $\vec{o_N} \in O_N$ and divide by the sum of $\rho_{\vec{o_S}}$ for all $\vec{o_S} \in O_S$.  This produces an ``outlier ratio'': $OR = \frac{\sum_{\vec{o_N} \in O_N} \rho_{\vec{o_N}}}{\sum_{\vec{o_S} \in O_S} \rho_{\vec{o_S}}}$    

Finally, we multiply the outlier ratio by $\text{cos}(\vec{\mu_S},\vec{\mu_N})$ (the cosine distance between the mean of the known samples and the mean of the new batch), divide that by $\text{cos}(\vec{\mu_S},\vec{\mu_S}+\vec{\sigma_S})$ (the cosine distance between the mean of the known samples and the mean of the known samples plus their standard deviation), and multiply that by the denominator of the outlier ratio.  This approximates how many times more dissimilar a given batch is from the mean of the known class than a random sample that falls within the vector subspace spanned by the known class samples would be.  Given that a new sample of a known class may not fall exactly within the vector subspace spanned by the samples in the data, we want this dissimilarity threshold to be greater than 1, acknowledging that the subspace defining a class may expand as new samples belonging to that class are encountered.  Therefore we define a dissimilarity threshold $T$, and if $\frac{OR \times \text{cos}(\vec{\mu_S},\vec{\mu_N})}{\text{cos}(\vec{\mu_S},\vec{\mu_S}+\vec{\sigma_S}) \times \sum_{\vec{o_S} \in O_S} \rho_{\vec{o_S}}} > T$ we say that the batch of new samples likely belongs to a class that is not one of the known classes.

\vspace*{-3mm}
\subsection{Results and Discussion}
\vspace*{-2mm}

\begin{figure}[h!]
    \centering
    \vspace*{-2mm}
    \includegraphics[width=.48\textwidth]{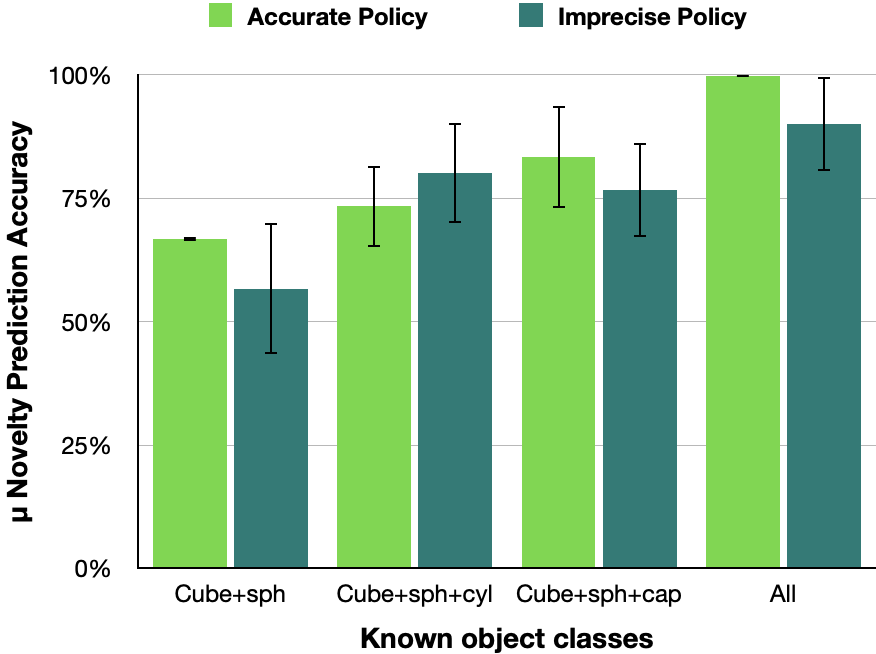}
    \includegraphics[width=.48\textwidth]{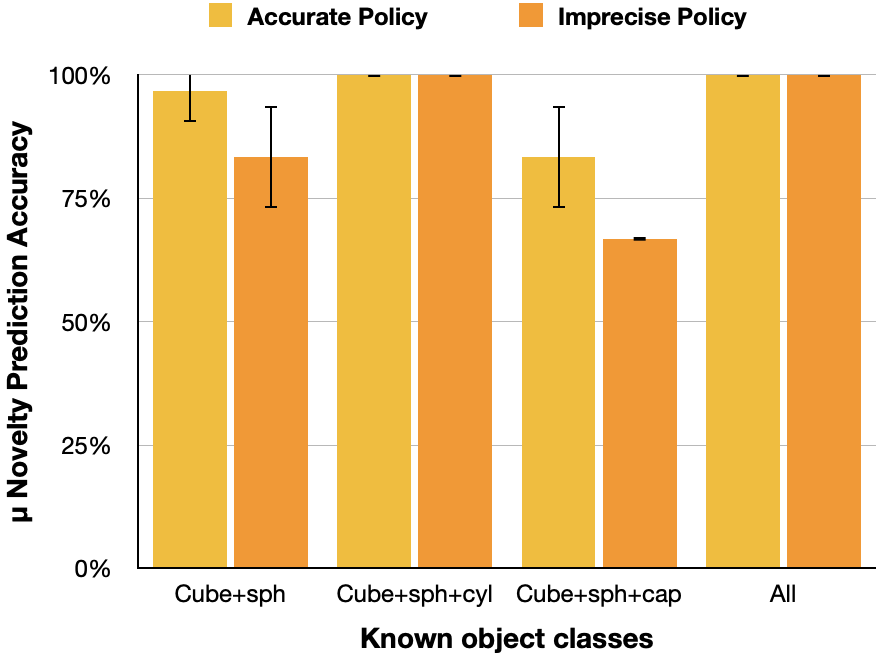}
    \vspace*{-2mm}
    \caption{Novel class identification accuracy under each condition. Left chart shows results without the VoxML-derived jitter information, and right chart shows results with it.}
    \vspace*{-3mm}
    \label{fig:novel-acc}
\end{figure}

We implemented tests where the classifier model was trained to classify different sets of known classes (cube and sphere, cube and sphere and cylinder, cube and sphere and capsule, and all four).  We also conducted an experiment where we trained the 1D CNN without the VoxML-derived jitter force feature, which implicitly encodes the axis of symmetry of the theme object, to compare what information this adds to the model.  We conducted 10 experiments under each condition with each dataset, using a dissimilarity threshold value of $T = 25$.  Correct results were identifying cylinder and capsule as novel classes where they were not already known, and not identifying small cube as a novel class.  Fig.~\ref{fig:novel-acc} shows the aggregate results with confidence intervals.

We can correctly identify the novelty of cylinders and capsules where appropriate, and identify batches of small cubes as instances of the known cube class, simply based on the way they behave in the stacking task.  The imprecise policy data is somewhat more challenging, because even stable objects like cubes fall off the bottom cube more often due to bad placement.

Using the VoxML-derived information in the object classifier boosts novel class detection performance up to 25\%.  Without it, the model cannot infer which axis the theme object moved along when the release jitter was applied, and so most cylinder embeddings end up nearly identical to cube embeddings.  Therefore it seems that this use of VoxML encodes information useful for common-sense reasoning a la \cite{hobbs1984sublanguage} into the model.

When cylinder is a known class to the model, it is much easier to identify capsule as a novel class than when the reverse is true, suggesting that order of class acquisition is important.  When the VoxML inputs are included in classifier training, our method identifies capsule as a novel class 100\% of the time, but when capsule was known first, cylinder was only classified as novel with 75\% accuracy or less.  We hypothesize that cylinder-to-cube is a much more fine-grained distinction than capsule-to-cube or capsule-to-sphere.  Because capsule is more obviously different in its behavior compared to cube or sphere, the capsule vectors take more ``space'' in the overall embedding space, making it hard to distinguish cylinder embeddings from other classes (usually cubes).  However, as in Sec.~\ref{sec:tl}, the source task starts with the two ``extremes'' {\it cube} and {\it sphere}, meaning that the geometric features relevant to detecting round and flat aspects of objects are already present.

\begin{wrapfigure}{l}{.71\textwidth}
    \centering
    \vspace*{-3mm}
    \includegraphics[width=.35\textwidth]{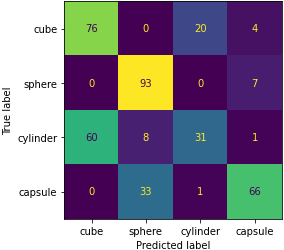}
    \includegraphics[width=.35\textwidth]{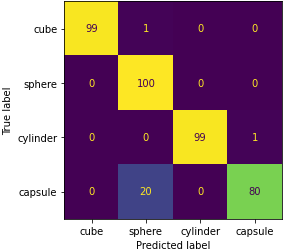}\\
    \includegraphics[width=.35\textwidth]{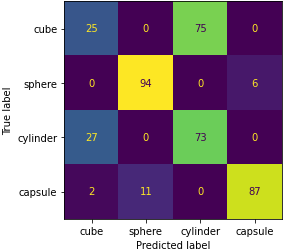}
    \includegraphics[width=.35\textwidth]{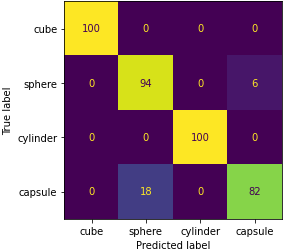}
    \vspace*{-3mm}
    \caption{Aggregated CNN outputs over the dev-test set.}
    \vspace*{-5mm}
    \label{fig:classifier-cms}
\end{wrapfigure}

Finally, even without the VoxML-derived information available to the model, the novel class detection method can successfully determine that the small cube is not a novel class, as shown by the right two bars in Fig.~\ref{fig:novel-acc}.  However, a deeper look into the classifier outputs show that although small cubes are being correctly labeled ``not novel,'' the most similar class they are subsumed into is often not {\it cube} but {\it cylinder}.  Fig.~\ref{fig:classifier-cms} shows the CNN classifier outputs over the 10 episode dev-test set, aggregated over all 10 novel concept detection experiments.  The two confusion matrices on the left show classification results without the VoxML-derived inputs.  The two on the right show results with those inputs.  The top two are from the {\it accurate policy} evaluation, and the bottom two are from the {\it imprecise policy} evaluation.  Without the VoxML-derived inputs, we see frequent confusion between sphere and capsule, and between cube and cylinder.  The VoxML-derived inputs are critical to correctly classifying the behavioral distinctions between objects like cubes and cylinders in the first place, in order to assess these classes' novelty.

The outputs of this procedure could be extended into the reinforcement learning itself, where if the policy learned on cubes fails in an environment containing a novel object, the policy itself is revised to allow the object to be stacked successfully (such as making sure all cylinders are placed upright). This is the subject of future work in this area.

\vspace*{-4mm}
\section{Conclusion}
\vspace*{-2mm}

Metacognition is defined as an actor's awareness and understanding of its own thought processes and the patterns underlying them.  More simply it may be called ``thinking about thinking'' \citep{blakey1990developing}. A metacognitive agent is one that, among other capacities, analyzes its own model of the environment and world, including the individual concepts it considers therein, figures out when it needs to be updated, how, and why.  In this paper we have demonstrated a certain subset of these capabilities in ML-based cognitive systems, in a domain intentionally reminiscent of geometric children's toys.

The neural network learns important distinguishing geometric features of the objects based on their behavior, and these features can be nonlinearly recombined to recognize new objects (this is demonstrated convincingly, but not necessarily conclusively, by adding new hidden layers). Embeddings based on these features can be used to detect outliers, and statistical techniques can be used to group outliers together into novel distinct classes.

Interestingly, in our work, none of the individual components---the neural network, the environment model, the analysis of embedding spaces, or the statistical metrics---can definitively be said to be solely responsible for the capabilities we demonstrate herein.  It is the hybrid approach and combination of techniques that led us to this result.  Ablation tests of the individual components to quantify the contribution of each is the topic of ongoing research.

The applicability of this same pipeline to other domains and shapes is to be determined, but the fact that distinctions can be made at the embedding level (even for objects unknown to the classifier) suggests that once a vector representation is acquired for an input sample of potentially any modality, similar assessments can be made over it.

The current work is limited primarily by the stacking task used to expose object properties, which in particular limits the number of objects with distinctions that can be revealed through this task.  The key features of flatness vs. roundness can be exposed through this task, as can other implicit distinctions such as length distinctions along an axis.  However other distinctions, such as those between irregular shapes, may require other tasks to expose.  These may be simple variants such as reversing the order of stacking and varying the destination object class instead of the theme.  But other, significantly different tasks may be required, such as fitting objects into the correctly shaped hole.  This is similar to how toddlers learn about different object properties through different activities.

Finally, unlike human learning, most modern AI techniques require very large volumes of data, long training times, or specialized (often expensive) hardware.  The models and processes we have developed here are small, lightweight, and can be trained or performed on as little as a laptop CPU, though GPU training provides a speed benefit for use in interactive systems.

\vspace*{-4mm}
\section{Future Work}
\vspace*{-2mm}

\begin{wrapfigure}{r}{.5\textwidth}
    \vspace*{-3mm}
    \centering
    \includegraphics[height=2in]{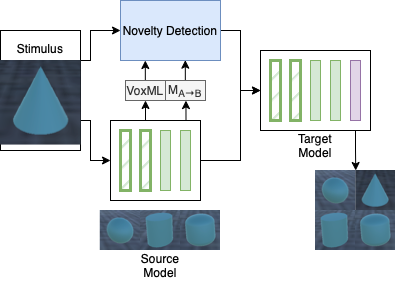}
    \vspace*{-3mm}
    \caption{Proposed integration architecture.}
    \label{fig:proposed-arch}
    \vspace*{-3mm}
\end{wrapfigure}

Particular points of future work in the line of research include estimating the dimensionality of the representations that are being transferred during the model expansion process.  Since we add one layer to detect each new object, this necessarily raises concerns about the scalability of the approach.  However, we surmise that the 25 dimensions used in the final layer of the transfer-learning neural network may not all be necessary to store all the information transferred between networks, or, if they are, then a larger embedding size may facilitate transfer learning between multiple types while not requiring the addition of a new layer each time.  In addition, we plan to investigate the number of layers that need to be frozen for successful transfer learning (e.g., is freezing one layer enough, or would freezing three be better, or is the number of layers perhaps a function of the number of classes or the domain?).  We also plan to introduce more shapes, including more irregular ones (e.g., a rock crystal).  However, with the introduction of new shapes comes a need for new tasks besides stacking to expose the distinctions between them.

The two suites of experiments herein were conducted separately, though using data from the same environment in variants of the same task. The primary goal of future work in this line of research is to integrate the two methods. This involves using the novelty detection method to provide the transfer learning model with an object that satisfies the detected set of novel properties such that the transfer learning model can be updated to achieve better accuracy with the new object set.  Since these currently are trained using different networks, the representations from the two networks need to be unified, either through a symbolic layer like VoxML, or by creating a mapping matrix $M$ between the two models $A$ and $B$ by solving for an affine transformation between paired embeddings \citep{mcneely2020exploring}.  Fig.~\ref{fig:proposed-arch} shows this proposed architecture. This integration facilitates the execution of both novelty detection and transfer learning in real time for use in interactive systems such as \cite{scheutz2019overview} or \cite{krishnaswamy2022affordance}.

\begin{acknowledgements} 
\noindent
We would like to thank our reviewers for their helpful comments.  This work was supported in part by a consultancy under National Science Foundation (NSF) award number CNS 2033932.  The views expressed herein do not reflect those of the National Science Foundation or the U.S. Government.  
%Please place your acknowledgements in an unnumbered section at the
%end of the paper. Typically, this will include thanks to reviewers
%who gave useful comments, to colleagues who contributed to the ideas, 
%and to funding agencies or corporate sponsors that provided financial 
%support.
\end{acknowledgements} 

\vspace*{-4mm}

{\parindent -10pt\leftskip 10pt\noindent
\bibliographystyle{cogsysapa}
\bibliography{format}

}

% Leave a blank line before the closing brace to ensure the final 
% reference has the proper indentation. 

\end{document}

%% file: cup-macros.tex
{\obeyspaces\gdef {\ }}
\global\newbox\codebox
\global\newbox\savedcodebox
\gdef\sverbatim{\bgroup\def\endsverbatim{\egroup\egroup\egroup\mbox{\box\codebo\
x}}\def\savecode{\egroup\egroup\egroup\global\setbox\savedcodebox\copy\codebox}\
\def\par{\egroup\vspace{-0.3em}\hbox\bgroup}\tt\obeylines\obeyspaces\global\set\
box\codebox\vbox\bgroup\hbox\bgroup}

\newcommand{\avmplus}[1]{{\setlength{\arraycolsep}{0.8mm}
                       \renewcommand{\arraystretch}{1.2}
                       \left[
                       \begin{array}{l}
                       \\[-2mm] #1 \\[-2mm] \\
                       \end{array}
                       \right]
                    }}
\newcommand{\att}[1]{{\mbox{\small {\bf #1}}}}
\newcommand{\attval}[2]{{\mbox{\small {\sc #1}}\ =\ {{#2}}}}

\newcommand{\attvaltyp}[2]{{\mbox{\small{\sc #1}}\ =\ {\myvaluebold{#2}}}}

\newcommand{\myvaluebold}[1]{{\mbox{\small {\bf #1}}}}

\newenvironment{enumerate*}%
  {\begin{enumerate}%
    \setlength{\itemsep}{0pt}%
    \setlength{\parskip}{0pt}}%
  {\end{enumerate}}

\newenvironment{itemize*}%
  {\begin{itemize}%
    \setlength{\itemsep}{0pt}%
    \setlength{\parskip}{0pt}}%
  {\end{itemize}}